\documentclass[12pt]{spieman}  % 12pt font required by SPIE;
\usepackage{amsmath,amsfonts,amssymb}
\usepackage{graphicx}
\usepackage{setspace}
\usepackage{tocloft}
\usepackage{lineno}
\usepackage{float}
\usepackage{multirow}
\usepackage{caption}
\usepackage{subcaption}
\usepackage{comment}
\usepackage{soul}
\usepackage{xcolor} % for custom colors
\usepackage{booktabs}
\usepackage{titlesec}

\titleformat{\paragraph}
{\normalfont\normalsize\bfseries}{\theparagraph}{1em}{}
\titlespacing*{\paragraph}
{0pt}{3.25ex plus 1ex minus .2ex}{1.5ex plus .2ex}

\DeclareUnicodeCharacter{2248}{\ensuremath{\approx}}

\title{Evaluating Federated Learning approaches for mammography under breast density heterogeneity}

\author[a,b,$*^{1}$]{Gonzalo Iñaki Quintana}
\author[a,c,d,$*^{2}$]{Franco Martin Di Maria}
\author[a,$*^{3}$]{Laurence Vancamberg}

\affil[a]{GE HealthCare, 283 Rue de la Minière, 78530 Buc, France}
\affil[b]{ENS Paris-Saclay, 4 Av. des Sciences, 91190 Gif-sur-Yvette, France}
\affil[c]{IMT Atlantique, 655 Av. du Technopôle, 29280 Plouzané, France}
\affil[d]{Universidad de Buenos Aires, Viamonte 430, C1053 CABA, Argentina}

\cftpagenumbersoff{figure}
\cftpagenumbersoff{table} 
\begin{document} 
\maketitle

\begin{abstract}

\textbf{Purpose:} Breast density is a key factor that influences mammography interpretation and is a major source of heterogeneity in multicenter datasets. Such heterogeneity poses challenges for collaborative machine learning across institutions, particularly in Federated Learning (FL). This study aims to evaluate the impact of breast density–induced heterogeneity on FL for mammography image classification and to assess the robustness of common FL algorithms in realistic clinical settings.

\textbf{Approach:} We conducted experiments on mammography datasets under two scenarios: (1) a strongly heterogeneous setting where each participating site contributed exclusively low- or high-density cases, based on the BI-RADS density score, and (2) a population-based setting simulating breast density distributions observed in White and Asian populations. For the strongly heterogeneous setting, we evaluated two configurations: one with 2 clients, where the cases were grouped as BIRADS A–B and C–D, and one with 4 clients, where each site contained cases of a single BIRADS density. We compared three FL methods (FedAvg, FedProx, SCAFFOLD) against centralized training, local-only training, and naïve aggregation approaches, including model Ensembling and weight averaging.

\textbf{Results:} Across both scenarios, FL consistently achieved performance comparable to centralized training, while local models and naïve aggregation approaches underperformed in the presence of strong heterogeneity. Notably, FedAvg achieved accuracy on par with or exceeding centralized training, demonstrating resilience to breast density–induced data imbalance without requiring specialized heterogeneity mitigation algorithms.

\textbf{Conclusions:} These findings show that FL can effectively address breast density–related heterogeneity, supporting its feasibility for real-world mammography workflows. The demonstrated robustness of FedAvg underscores the potential for broad clinical deployment of FL, enabling collaborative model development while maintaining data privacy.

\end{abstract}

% Include a list of up to six keywords after the abstract

\keywords{Federated Learning, Deep Learning, Computer Aided Detection, mammography, breast density}

% Include email contact information for corresponding author

{\noindent \footnotesize\textbf{$*^{1}$} {Gonzalo Iñaki Quintana,  \linkable{gonzaloinaki.quintana@gehealthcare.com}}}

{\noindent \footnotesize\textbf{$*^{2}$} {Franco Martin Di Maria,  \linkable{francomartin.dimaria@gehealthcare.com}}}

{\noindent \footnotesize\textbf{$*^{3}$} {Laurence Vancamberg,  \linkable{laurence.vancamberg@gehealthcare.com}}}

\begin{spacing}{2}   % use double spacing for rest of manuscript

\section{Introduction}

Breast cancer is among the most prevalent cancers affecting women. In 2020, it was the most commonly diagnosed cancer worldwide, representing 11.7\% of all cancer cases (approximately 2.3 million) and accounting for 6.9\% of all cancer-related deaths \cite{cancer_statistics_2020}. Research indicates that early detection significantly improves survival rates, with up to 90\% of patients being curable when diagnosed at an early stage \cite{cancer_statistics_2016_miller}. X-ray imaging modalities such as Full Field Digital Mammography (FFDM) and Digital Breast Tomosynthesis (DBT) are essential tools for the detection, diagnosis, and monitoring of breast cancer.

Deep learning-based Computer-Aided Detection and Diagnosis (CAD) systems are designed to support radiologists in identifying and interpreting radiological findings, by performing classification, lesion detection or segmentation tasks. However, developing these models demands access to large, well-annotated datasets that reflect the variability across populations, imaging devices, and post-processing techniques. Gathering such data from multiple clinical sites is often complicated by legal and practical constraints related to health data privacy regulations, which pose a significant barrier to the advancement of DL-based CAD systems for mammography.

Federated Learning (FL) offers a promising alternative by enabling model training with data from different clinical sites without the need to collect and store data in a centralized location. This approach can help address one of the major limitations in leveraging diverse datasets from different clinical environments. While FL has demonstrated encouraging performance in settings where data across sites is relatively homogeneous, its effectiveness decreases significantly in heterogeneous settings—where data distributions differ between clients or institutions. This limitation is especially critical in mammography, where substantial inter-site variability arises from differences in patient demographics, imaging hardware, vendor-specific acquisition protocols, and post-processing algorithms, resulting in complex and highly non-identically distributed data.

In this work, we study the effect of data heterogeneity on the training of Federated Learning models for mammography. In particular, we focus on the heterogeneity resulting from variations in breast density across different populations encountered at the clinical sites, and study its impact in classification models. To the best of our knowledge, this is the first study of this type.

\subsection{Breast density}

Breast density refers to the proportion of fibroglandular (dense) tissue relative to fatty (non-dense) tissue in the breast, as observed on a mammogram. It is a clinically significant attribute, as dense tissue appears white on mammograms—similar to tumors—thereby reducing the sensitivity of cancer detection \cite{boyd2007mammographic, Nazari2018}. Breast density is largely hereditary \cite{boyd2007mammographic, ursin2009relative} but it is also influenced by other factors, such as menopausal status, body weight, parity status, and number of births \cite{greendale2003, li2005association, Nazari2018}.

\begin{sloppypar}
Population differences also contribute to variations in breast density. A study by Kerlikowske et al. \cite{kerlikowske2023impact} demonstrated that dense breasts are more prevalent among Asian women (66.0\%), followed by non-Hispanic White women (45.5\%), Hispanic/Latina women (45.3\%), and non-Hispanic Black women (37.0\%). These demographic disparities further complicate the distribution of breast density across clinical sites, amplifying heterogeneity in federated settings.    
\end{sloppypar}

\begin{figure}[tb]
    \centering
    \includegraphics[width=0.9\linewidth]{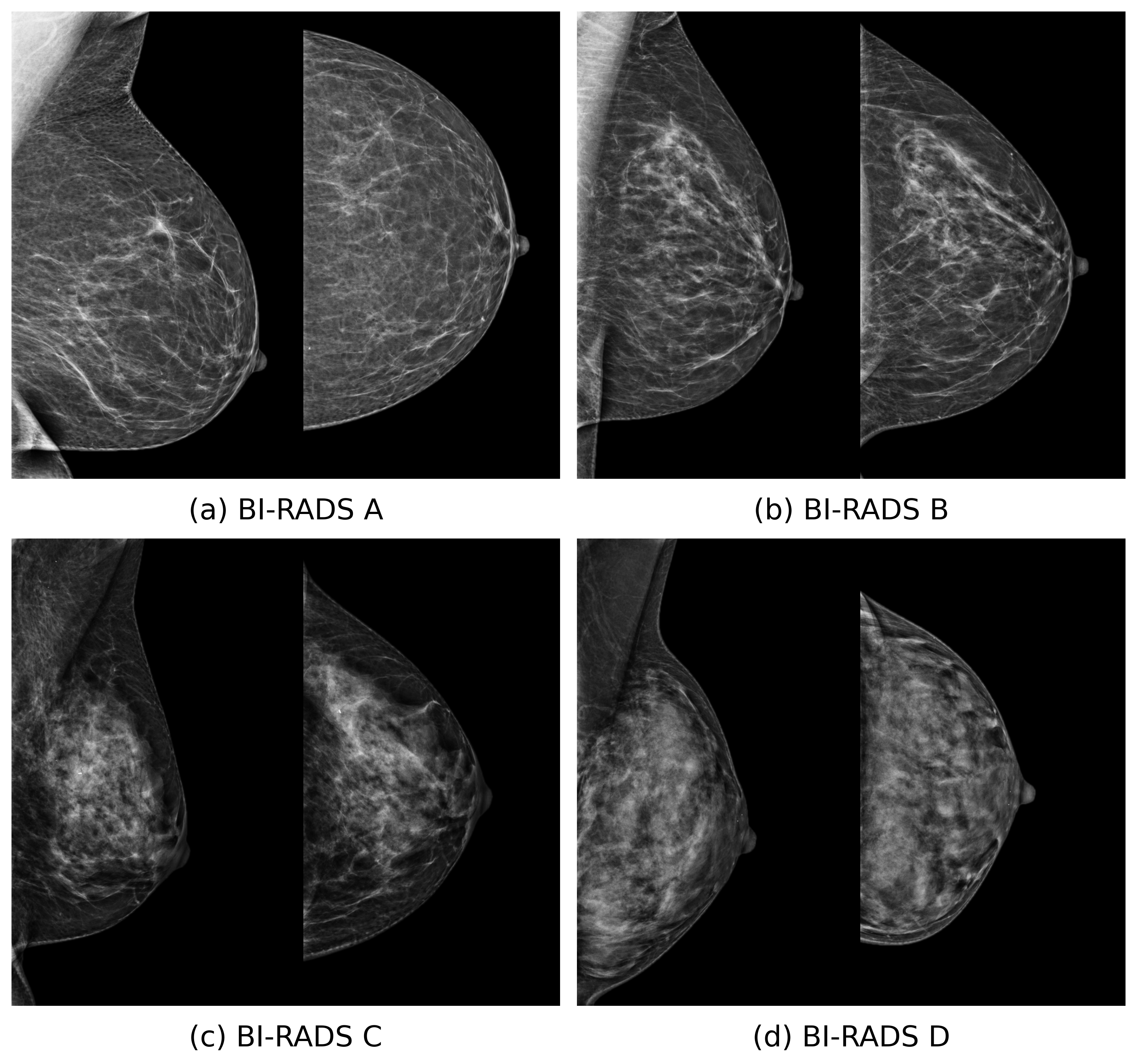}
    \caption{BI-RADS Breast composition categories. Each subfigure shows two views of the same breast: MLO on the left and CC on the right.}
    \label{fig:IntroBIRADSCategories}
\end{figure}

To standardize the assessment of breast density, radiologists often refer to the guidelines of the American College of Radiology (ACR) in the Breast Imaging Reporting and Data System (BI-RADS) \cite{BI-RADS}, which classifies breast composition into four categories: (1) Almost entirely fatty (about 10\% of women), (2) Scattered areas of fibroglandular density (about 40\%), (3) Heterogeneously dense (about 40\%), and (4) Extremely dense (about 10\%). These categories not only impact radiological interpretation but also affect the underlying data distribution used to train machine learning models. Figure \ref{fig:IntroBIRADSCategories} depicts two example mammograms from each of the BI-RADS categories.

\subsection{Federated Learning}

\begin{figure*}[tb]
    \includegraphics[width=\linewidth]{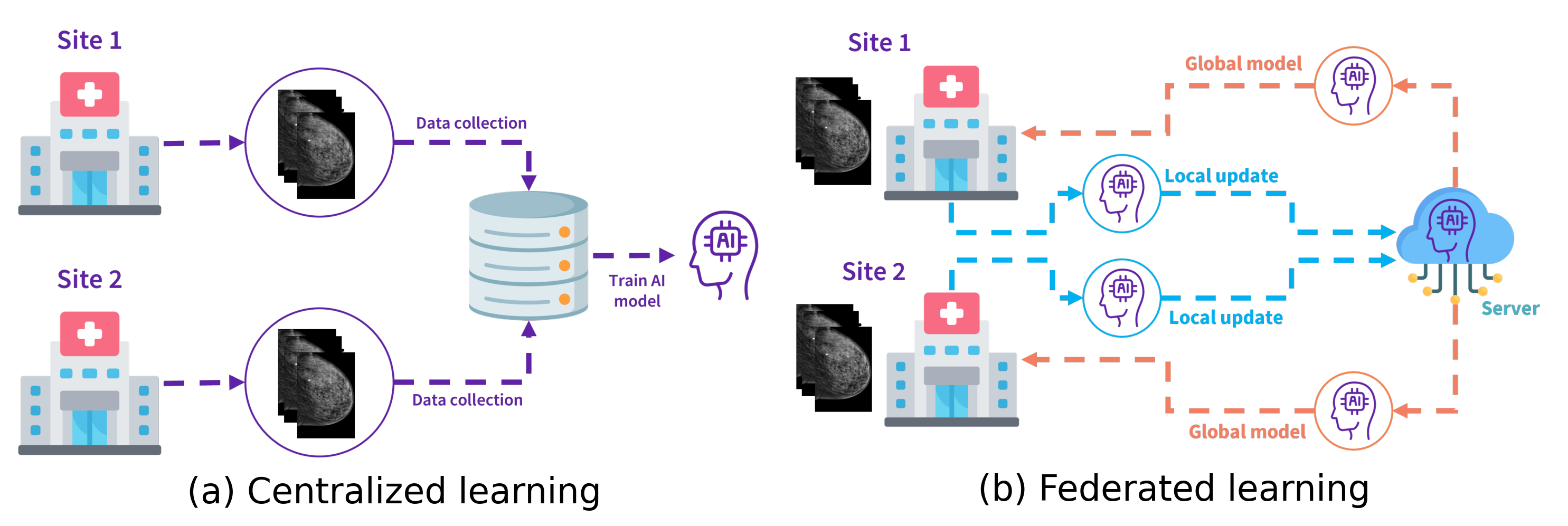}
    \caption{Centralized and federated learning paradigms. Icons from Flaticon.com}
    \label{fig:centralized_vs_fl}
\end{figure*}

Federated Learning (FL) is a decentralized approach to training machine learning models, where a group of clients — such as mobile devices, hospitals, or organizations — collaboratively learn a shared global model without exchanging raw data. This framework aims to increase the volume of training data available while preserving privacy and reducing the costs associated with data collection. Figure \ref{fig:centralized_vs_fl} contrasts the centralized learning paradigm to server-orchestrated Federated Learning, a variant of FL in which a central server coordinates the training process. In a standard centralized setting, data need to be collected from different clinical sites and stored in centralized databases before training Deep Learning models. In Federated Learning, model training is structured into multiple global rounds, during which multiple clients and a central server collaboratively update a shared model without exchanging raw data. At the beginning of each global round, the server distributes the current global model weights to all the participating clients. Each client then updates the model locally using its private data for a fixed number of local iterations and sends the updated model back to the server. The server aggregates these updates to form the next version of the global model. This cycle is repeated for several global rounds until convergence. In real-world applications, clients often vary significantly in their computational resources and data distributions, and communication between them is usually limited or expensive.

\section{Related Works}
\label{sec:intro_related_works}

\subsection{Mammography image classification}

There are two primary approaches for classifying 2D mammography images: full image-based methods and patch-based methods \cite{shen2019deep, quintana2023mammo}.

Full image-based methods involve using a Deep Convolutional Neural Network (DCNN) that processes the entire mammogram to produce a class probability or cancer score. These methods typically adapt models originally designed for natural images and rely solely on image-level annotations—ground-truth labels assigned to the entire image. For instance, Ayana et al. proposed a patch-less, multi-stage transfer learning approach that first fine-tunes an EfficientNetB2, pre-trained on ImageNet, on a dataset of cell line images, and subsequently on full mammogram images for breast mass classification \cite{ayana2022patchless}. However, mammograms present a unique challenge: the regions containing lesions—which are most informative for classification—often occupy only a small fraction of the image. As a result, full image-based classifiers are difficult to train effectively and generally require large datasets to perform well \cite{shen2019deep, quintana2023mammo}.

\begin{sloppypar}
Patch-based methods, on the other hand, leverage lesion-level annotations by first training a CNN on image patches. This patch-level classifier is then extended to the full image, typically by applying it in a sliding window fashion across different regions of the mammogram and aggregating the predictions using techniques such as voting or averaging \cite{patch_clsf_sliding_wdw, patch_clsf_sliding_wdw2, patch_clsf_sliding_wdw3, patch_clsf_sliding_wdw4}. For example, Terrassin et al. adapted a U-Net3+ architecture for mammogram classification and segmentation by decomposing the images into patches processed sequentially \cite{terrassin2024weakly}. A key limitation of this approach is that each patch is treated independently, which prevents the model from capturing spatial relationships between patches.    
\end{sloppypar}

A more integrated method involves the application of a trained patch classifier on the mammogram and then fine-tuning additional convolutional layers to aggregate features from different regions \cite{shen2019deep, petrini_2021, quintana2023mammo}. This approach enables efficient training of full image classifiers that incorporate lesion-level information, while avoiding the redundancy and inefficiencies of sliding window techniques.

\subsection{Impact of breast density}

Breast density is a well-documented factor that affects the accuracy of mammographic breast cancer detection. On mammographic imaging, dense breast tissue exhibits high radiopacity, which impairs the conspicuity of cancerous lesions and complicates their differentiation from adjacent parenchymal structures \cite{Nazari2018}. As a result, women with dense breasts face a higher likelihood of both false negatives—where cancer goes undetected—and false positives, which can lead to unnecessary follow-up procedures.

Empirical studies support these concerns. Von Euler-Chelpin et al. observed that mammography sensitivity decreased as breast density increased in Danish women \cite{vonEulerChelpin2019}: from about 80\% in BI-RADS category 1 (fatty breasts) to approximately 75\% in category 2 (scattered density), 69\% in category 3 (heterogeneously dense breasts), and 47\% in category 4  (extremely dense breasts). Similarly, Van der Waal et al. reported a drop in sensitivity from 75.7\% in fatty breasts to 57.8\% in dense breasts \cite{vanderWaal2017}. These findings illustrate how breast density can significantly impair cancer detection performance in clinical settings.

Deep learning models for mammogram analysis exhibit similar trends. Suh et al. trained a DenseNet-169-based model and found that its AUC-ROC dropped from 0.984 for BI-RADS A (lowest density) to 0.902 for BI-RADS D (highest density) \cite{Suh2020}. Omoleye et al. further examined this issue using Mirai \cite{Yala2021}, a state-of-the-art risk prediction model, and found that its one-year risk prediction AUC-ROCs were 0.72 for non-dense and only 0.58 for dense breasts \cite{Omoleye2023}. These results suggest that model performance deteriorates significantly as breast density increases.

There is ongoing debate about whether models should be trained on datasets that include all breast densities or on subsets of interest. For example, Kim et al. trained models for breast cancer risk prediction in Asian women and found that a model trained only on dense mammograms outperformed another model trained on a mixed-density dataset in most metrics—except sensitivity \cite{Kim2023}. However, since each model was tested on a dataset that matched its training distribution, direct comparison is problematic.

Interestingly, commercial CAD systems show different behavior. Brem et al. evaluated the FDA-approved Second Look system and found no significant performance difference between dense and non-dense breast images \cite{Brem2005}. Dustler et al. \cite{Dustler2020} and Riveira-Martin et al. \cite{RiveiraMartin2023} reported similar results with different versions of the Transpara system, finding no statistically significant differences in AUC-ROCs across BI-RADS density categories. These discrepancies with open-source models likely stem from differences in dataset scale, curation, and model optimization. Commercial systems are typically trained on much larger and more diverse datasets and undergo rigorous validation and regulatory approval, which may contribute to their robustness. This underscores the importance of dataset diversity and thoughtful model design in addressing breast density-related challenges in mammography.

\subsection{Federated Learning}

Recent research in Federated Learning for brain lesion segmentation has shown performance on par with centralized training and superior to training on isolated sites \cite{sheller2019multi}. In mammography, FL has also shown promising results on image classification for cancer detection \cite{FL-curriculum-learning-MI}, on breast density classification \cite{2020_FL_breast_density_classification}, and on continuous breast density prediction \cite{muthukrishnan2022mammofl}. Jiménez-Sánchez et al. \cite{FL-curriculum-learning-MI} studied the impact of multi-vendor FL training for malignant-benign image classification and proposed FedAlign, a memory-aware curriculum learning method  for improving model consistency across clients. Roth et al. \cite{2020_FL_breast_density_classification} trained a four-class breast density classification model, corresponding to the four BI-RADS categories, in a real-world federated learning setting involving seven institutions. For predicting a continuos breast density value (as the proportion of fibroglandular tissue relative to fatty tissue), Muthukrishnan et al. applied FL to train a double U-Net on multi-institutional data, showing that its performance and generalizability nearly matched those obtained with centralized training on the same datasets \cite{muthukrishnan2022mammofl}. Despite these promising results, Federated Learning algorithms face the challenge of identically distributed (non-IID) data that are not independent of each other, resulting from statistical heterogeneity between institutions and patient populations \cite{konevcny2016federated}. Studies have shown that common optimization methods, such as Stochastic Gradient Descent, perform poorly under such heterogeneity \cite{konevcny2016federated, zhao2018federated}. Furthermore, clinical data imbalance, such as variations in disease prevalence, demographics, and risk factors, further complicates the aggregation of local model updates \cite{rieke2020future}. Most of the studies on Federated Learning in the context of medical imaging and mammmography are based on the Federated Averaging (FedAvg) algorithm \cite{mcmahan2017communication}, the most commonly used and standard approach for server-orchestrated FL. In FedAvg, local updates are performed using SGD, making it a natural extension of centralized SGD to the federated setting.

The slow and unstable convergence of FedAvg in heterogeneous settings is often attributed to client drift \cite{li2020fedprox, karimireddy2021scaffold}, a phenomenon that arises when the optimal model parameters for each client's local data differ from the global optimum. This discrepancy causes local updates to deviate from the global learning objective, particularly under non-IID data distributions.

To address client drift, we consider two algorithms proposed in previous works: FedProx \cite{li2020fedprox} and SCAFFOLD \cite{karimireddy2021scaffold}. FedProx modifies the local optimization problem by adding a proximal term that penalizes large deviations from the global model, thereby constraining local updates to remain closer to the global direction. In contrast, SCAFFOLD introduces a variance reduction mechanism that estimates and corrects client drift using control variates maintained at both the client and the server. These control variates are used during local updates to counteract the bias introduced by data heterogeneity.

\subsection{Model Ensembling and Weight Averaging}

Model Ensembling is a well-established technique in machine learning for improving generalization by combining the predictions of multiple models \cite{geurts2002contributions, goodfellow2016deep, wood2023unified}. A related approach, weight averaging, instead combines the model parameters themselves rather than their outputs. Techniques like Stochastic Weight Averaging (SWA) \cite{izmailov2018averaging} and those that explore the loss surface geometry \cite{garipov2018loss} demonstrate that averaging weights from models converged to different points in the loss landscape can yield a single model with improved generalization.

Model Soup \cite{wortsman2022model} is a recent extension of this idea. Instead of averaging weights along a training trajectory (as in SWA), Model Soup averages the parameters of independently trained models, selected based on validation performance. This can result in a single model that performs better than any individual model in the soup, without requiring an ensemble at inference time.

We argue that federated learning (FL), and in particular the FedAvg algorithm, can be interpreted as a dynamic and distributed form of weight averaging, closely related to the concept of Model Soup. In FedAvg, each global round consists of local training on heterogeneous clients followed by weight averaging across clients. If the number of local steps is sufficiently large, each global step effectively corresponds to generating diverse models trained on different data distributions. The subsequent aggregation step can be seen as averaging these models in a manner analogous to Model Soup. Furthermore, by running multiple global rounds and evaluating the resulting global models, FedAvg implicitly performs a form of model selection over these averaged weights—akin to choosing the best combination of models in a soup.

\section{Materials and methods}

\subsection{Datasets}
\label{materials_methods:dataset}

In this study, all the experiments were conducted in the GE HealthCare (GEHC) dataset, an internal set of mammograms generated by GE Senograph Pristina \texttrademark{}. All data was collected anonymously from a single institution following the EU General Data Protection Regulation. It consists of 1,294 patients: 207 cases of mass and/or calcification diagnosed with cancer through biopsy (malignant), 318 diagnosed with benign lesions through biopsy and 769 considered normal, meaning they do not have any benign or malignant lesions. While all malignant lesions were masses and calcifications, 37 benign lesion belonged to less common categories, such as architectural distortions. The mammograms of normal patients were confirmed by a follow-up exam. The GEHC dataset includes bounding box annotations around the lesions, specifying its lesions type (mass, calcification, other lesions) and its pathology (benign or malignant), and the image-wise BIRADS density score (which was found to be consistent across different images from the same patient).

The GEHC dataset was split into a development (80\%) and a test (20\%) subsets in a stratified fashion, considering lesion type, pathology, and density. The split was conducted at a patient level by assigning each patient a lesion type, lesion sub-type, lesion pathology (benign or malignant), and breast density from all its mammograms. A priority order was defined for selecting the lesion type and class for each patient, giving highest priority to malignant lesions, followed by benign masses or calcifications, and finally by less representative anomalies. In case of tie, the least representative lesion class of the so far processed dataset was chosen. Thus, if one patient had malignant and benign lesions, its split label would be that of the  malignant one, e.g., malignant spiculated mass. Only patients with all images labeled as normal were considered as normal. The development subset was then split across the clients in a stratified way, as detailed in \ref{materials_methods:fed_ler_setting}. The data of each client was subsequently split into training and validation subsets using 5-fold cross validation, with the same stratification as described above. All models were evaluated on the same test set.

\subsubsection{Patch extraction} 
\label{materials_methods:patch_extraction}

Random patches of fixed size were extracted from each mammogram of the dataset. The patch size was set to $256 \times 256$, as its classification performance was shown to be on par with larger patch sizes like $512 \times 512$, while requiring significantly fewer computational resources \cite{quintana2023mammo}. One normal patch, i.e., a patch containing only breast tissue and no lesions, was extracted from each mammogram by centering it in randomly selected locations of the breast. If a lesion was present in the image (considering only masses and calcifications), five additional random patches were sampled, each centered at a randomly selected location within the lesion. To ensure patch diversity, patches with an Intersection over Union (IoU) greater than 50\% with previously selected patches were discarded. Each resulting patch was associated to one label, given by its lesion type and pathology: NM (normal), BM (benign mass), BC (benign calcification), MM (malignant mass), and MC (malignant calcification). Overall, 7825 patches were sampled in the development set: 2700 NM, 1600 BM, 950 BC, 1875 MM, and 700 MC. In the test set, 3015 patches were sampled: 640 NM, 690 BM, 685 BC, 670 MM, and 330 MC. 

\subsection{Deep Learning model}

\begin{figure*}[tb]
\centering
\includegraphics[width=0.8\linewidth]{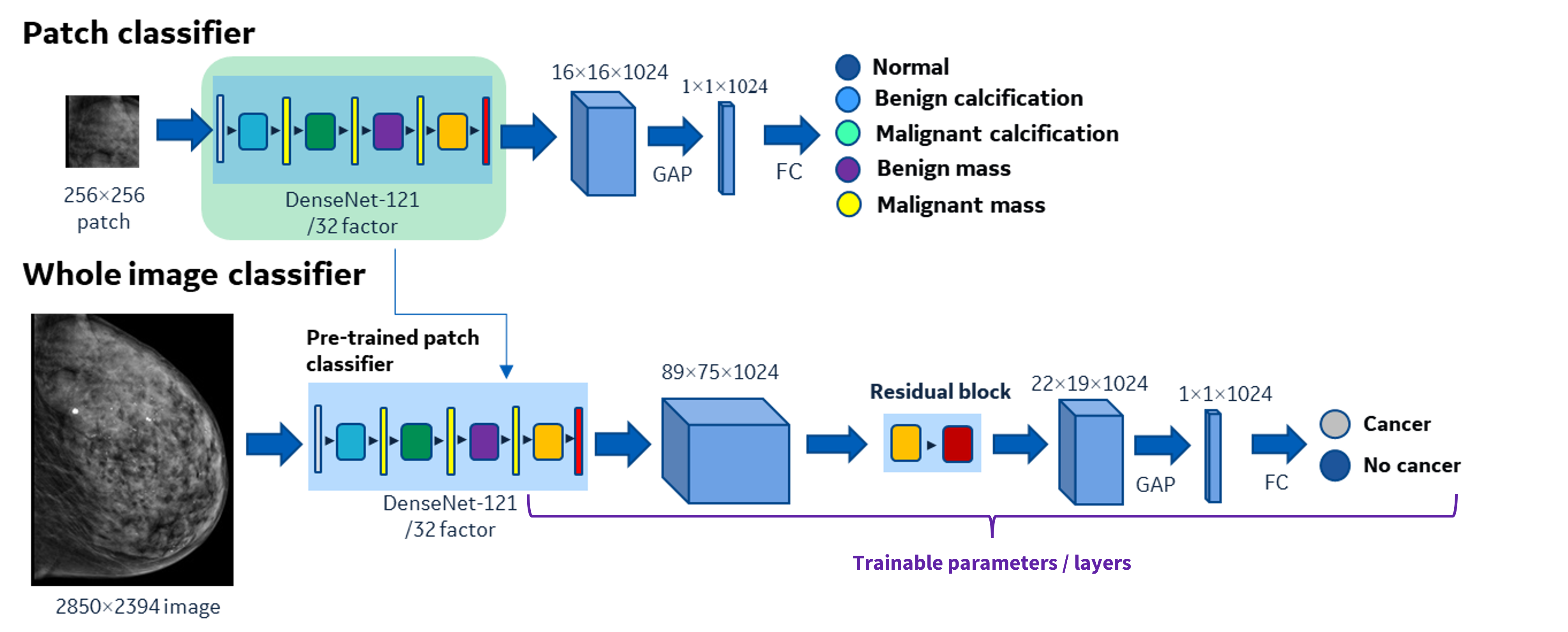}
\caption{Architecture of the patch-classifier (top) and whole image classifier (bottom), based on DenseNet-121. GAP: Global Average Pooling, FC: Fully Connected layer.}
\label{fig:model}
\end{figure*}

We adopted a patch-based approach for whole-image classification, as illustrated in Figure~\ref{fig:model}. First, a DenseNet-121 was trained to classify $256 \times 256$ image patches into five categories: normal, benign/malignant calcification, and benign/malignant mass (Figure~\ref{fig:model} – top). DenseNet-121 was selected over ResNet variants for its favorable balance of depth, parameter efficiency (7M vs. 21.3M in ResNet-34), and performance. The model was initialized with ImageNet weights and first trained on CBIS-DDSM \cite{cbis-ddsm} - the largest publicly available dataset of mammograms with lesion-level annotations - following previous work \cite{quintana2023mammo}. It was then fine-tuned on patches from the GEHC dataset.

To classify full mammograms (Figure~\ref{fig:model} – bottom), the classification head of the patch model was removed, and its feature maps were passed through a convolutional residual block and a new classification head for binary cancer prediction. To reduce GPU usage, only the final dense block and subsequent layers were trained on full mammograms. To further limit GPU memory usage, we employed a memory-efficient implementation of DenseNet-121 training \cite{pleiss2017memoryefficientdensenet}.

\subsection{Federated Learning setting}
\label{materials_methods:fed_ler_setting}

Two distinct federated settings with two clients were considered, with heterogeneity due to differences in the breast density distributions. These settings were formed by splitting the development data subset into the two clients, at a patient level. In the \textit{strongly heterogeneous} setting, one client contained exclusively data from patients with low breast density (BIRADS density scores A and B), while the other client contained data from high breast density patients (BIRADS scores C and D). In the \textit{population-based heterogeneous} setting, the data of the two clients was set to simulate real observed population differences regarding breast density. Following the study of \cite{BreastDtyRace}, patients were split between the two sites to match the breast density distributions of white and asian populations. We stress that this split was conducted solely using BIRADS density score, and no race or ethnicity information from the patients was used.

An additional strongly heterogeneous dataset for a 4 clients setting is created. In this dataset, each client contains samples for a single BIRADS density category, i.e., A, B, C, D. Due to the increased computational needs of this setting, in which 4 different GPUs are needed for each training, experiments in this setting are solely performed for the whole image classifier.

In all the settings, the splits between clients were conducted in a stratified fashion considering the type and sub-type of the lesions contained as well as their pathology, with the same priority system as in Section \ref{materials_methods:dataset}, keeping the number of patients in each client approximately the same. This enables the isolation of the effect of breast density heterogeneity.

To train the patch-classifiers, patches were extracted following the procedure described in \ref{materials_methods:patch_extraction} for each client at each split. We highlight that the BIRADS score is a measure of the density of the entire breast, and not of all the patches that can be extracted from the mammogram. Thus, a patch being labeled with BIRADS score C only implies that the image from which it was sampled has a density score C.

\subsection{Training strategies}
\label{materials_methods:training_strategies}

\begin{sloppypar}
Three main training strategies were considered: local training, centralized training and federated training. Local training refers to training a classifier on the local dataset of a specific client in the federated setting. Centralized training refers to training a classifier using the data from both federated clients, combined into a single dataset. Finally, federated training refers to training a classifier using the data from both clients in a distributed manner. Patch-classifiers and whole image classifiers were trained using these three strategies, both in the strongly heterogeneous and population-based heterogeneous settings. In the population-based heterogeneous setting, models were trained with the FedAvg algorithm. In the strongly heterogeneous setting, FedProx and SCAFFOLD were used to alleviate the impact of heterogeneity, and their results were compared to those of FedAvg. Additionally, the training strategies were compared to model Ensembling and Model Soup of the local models. The Ensembling results were obtained by averaging the logits of each one of the local models (two in this case) before applying the Softmax layers. The Model Soup consists of averaging the weights of the local models to create a new model for prediction. We highlight that while it has the same inference computational complexity as the other models, its inference complexity scales linearly with the number of clients.
\end{sloppypar}

\subsection{Evaluation}
\label{materials_methods:evaluation}

All the models from the strongly heterogeneous setting were evaluated on the original test subset, obtained from the development-test split of Section \ref{materials_methods:dataset}. For evaluating the population-based heterogeneity models, White and Asian population test subsets were sampled from the original test subset in the same stratified fashion as the client splits, to match the distributions described by del Carmen et al. \cite{BreastDtyRace}. The test subset that results of merging these two subsets is denoted as \textit{population-based} subset. It should be noted that this test subset contains fewer elements than the original test subset, which was also used to evaluate the population-based models. For each model and training setting, the 5-fold cross-validation (CV) results were aggregated. Additionally, for each fold, 100 bootstrap iterations were performed for calculating error bars and p-values. In both cases, the mean and standard deviation (assuming Gaussian errors) are reported. Statistical significance was assessed using the Wilcoxon Signed-Rank Test for paired comparisons on a shared test set. Patch-classifiers are compared in terms of the multi-class Accuracy and multi-class AUC-ROC-based metrics: One-vs-one AUC-ROC (AUC-ROC OvO) and One-vs-rest AUC-ROC (AUC-ROC OvR). The whole image (binary) classifiers are compared in terms of the ROC curves and AUC-ROC.

\section{Results}

\subsection{Population-based heterogeneous setting - 2 clients}
\label{sec:res_pop_based_het}

We first present and discuss the results of the training in the population-based heterogeneous setting, with the models evaluated in the original and population-based test sets.

\subsubsection{Patch-classifier}

\begin{figure}[h]
    \centering
    \includegraphics[width=0.75\linewidth]{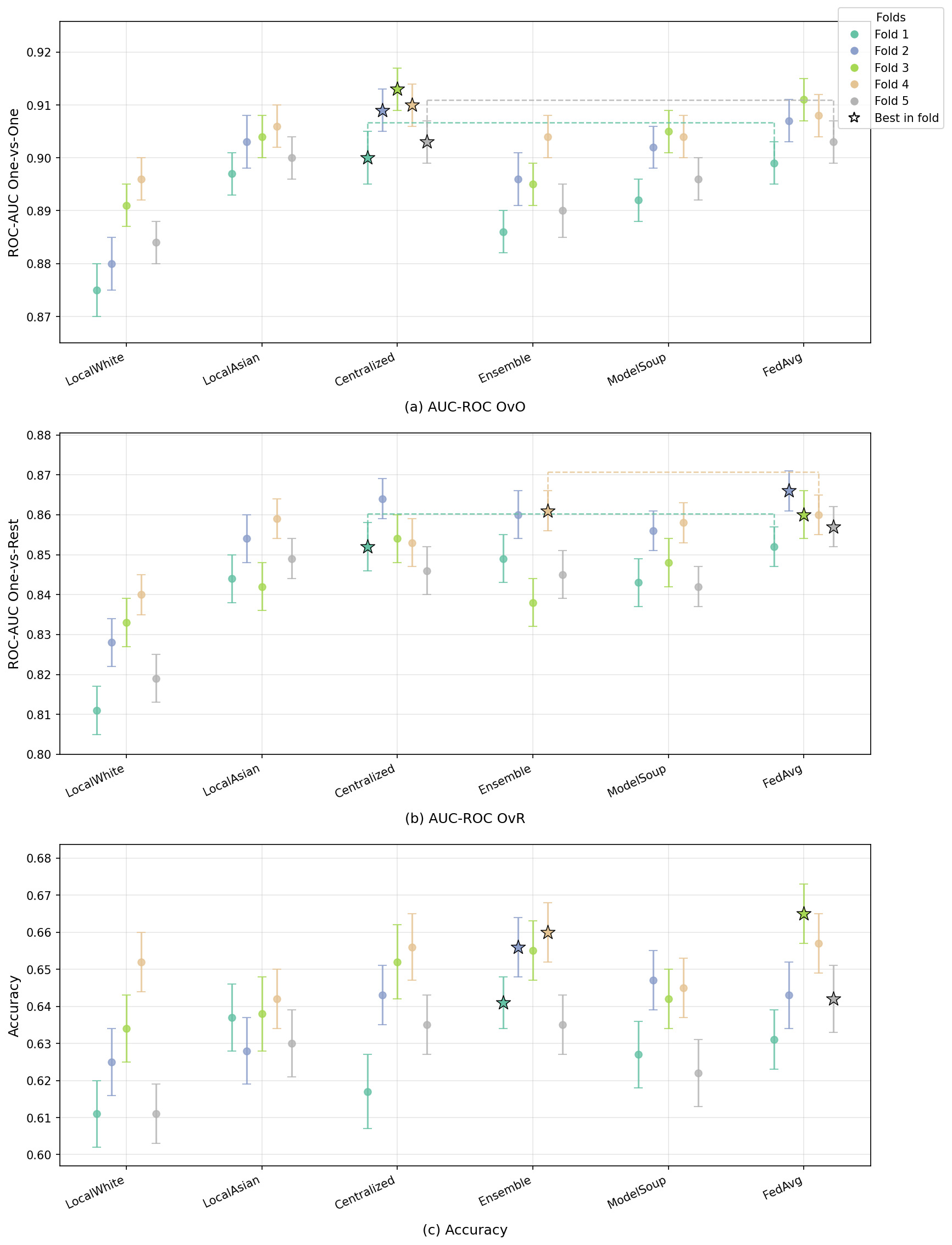}
	\caption{\small Folds bootstrap results of the patch-classifiers trained on the population-based heterogeneous setting, evaluated on the original test set (mean $\pm$ covariance). Best models for each fold are marked with a star symbol. Models that are not statistically different from the best model according to the Wilcoxon signed-rank test (p-values $>$ 0.1) are indicated by horizontal dashed lines.}
\label{fig:realistc_patch_std_Full_scatter_pvalue_welch}
\end{figure}

Figure \ref{fig:realistc_patch_std_Full_scatter_pvalue_welch} shows the per-fold results on the original test set. The FL model, trained with standard FedAvg algorithm, is the only model to equal the performance of the centralized model (no significant statistical difference), in all the considered metrics (AUC OvO, AUC-ROC OvR, Accuracy), outperforming the remaining training strategies. The model aggregation strategies fail to reach the performance of the centralized training in all the metrics (Ensembling is outperformed in terms of AUC-ROC OvO, Model Soup in terms of all the metrics). Local training on the site with a white population achieves the lowest performance in the original test for all the folds. In contrast, local training on the Asian population achieves performance on par with the centralized model in terms of Accuracy and AUC-ROC OvO, even though it underperforms in terms of AUC-ROC OvR. Local training on the Asian population site achieves significantly higher performance than on the White population site, but it is still outperformed by both centralized and federated training.

\begin{sloppypar}
Table \ref{tab:res_patch_het_training_het_test} presents the performance results on the population-based test set, as well as on the Asian and White subsets. Once again, FedAvg is the only model that matches the performance of centralized training across all metrics and data subsets. While Ensembling and Model Soup achieve centralized-level performance on the overall population-based test set and the White subset (leftmost and rightmost columns), they fall short on the Asian subset, specifically in terms of OvO AUC. Finally, we note that the high variability in Accuracy prevents us from drawing definitive conclusions in this setting.
\end{sloppypar}

\begin{table}[h]
\centering
\caption{\small 5-Folds CV results of the patch-classifiers trained on the population-based heterogeneous setting and evaluated on the population-based test sets (mean $\pm$ covariance). Bolded values indicate the best-performing models, whose differences in performance are not statistically significant (Wilcoxon Signed Rank test p-values greater than 0.1).}
\caption*{\textbf{Population-based setting - Patch Clfs. - Performance}}
\resizebox{\textwidth}{!}{
\begin{tabular}{lccccccccc}
\toprule
Model       & \multicolumn{3}{c}{Population based}                                                 & \multicolumn{3}{c}{Asian population-based}                                           & \multicolumn{3}{c}{White population-based}                                           \\
\cmidrule(lr){2-4} \cmidrule(lr){5-7} \cmidrule(lr){8-10}
            & AUC OvO                    & AUC OvR                    & Accuracy                   & AUC OvO                    & AUC OvR                    & Accuracy                   & AUC OvO                    & AUC OvR                    & Accuracy                   \\
\midrule
LocalWhite  & 0.888 $\pm$ 0.008          & 0.825 $\pm$ 0.012          & 0.623 $\pm$ 0.016          & 0.899 $\pm$ 0.009          & 0.828 $\pm$ 0.015          & 0.622 $\pm$ 0.027          & 0.879 $\pm$ 0.008          & 0.822 $\pm$ 0.012          & \textbf{0.624 $\pm$ 0.006} \\
LocalAsian  & 0.899 $\pm$ 0.005          & 0.843 $\pm$ 0.007          & 0.632 $\pm$ 0.010          & 0.913 $\pm$ 0.009          & 0.844 $\pm$ 0.019          & \textbf{0.649 $\pm$ 0.013} & \textbf{0.888 $\pm$ 0.011} & \textbf{0.842 $\pm$ 0.010} & \textbf{0.615 $\pm$ 0.013} \\
Centralized & \textbf{0.905 $\pm$ 0.005} & \textbf{0.849 $\pm$ 0.007} & \textbf{0.635 $\pm$ 0.017} & \textbf{0.923 $\pm$ 0.006} & \textbf{0.859 $\pm$ 0.018} & 0.649 $\pm$ 0.020          & \textbf{0.890 $\pm$ 0.009} & \textbf{0.841 $\pm$ 0.009} & \textbf{0.621 $\pm$ 0.019} \\
Ensemble    & 0.894 $\pm$ 0.007          & \textbf{0.846 $\pm$ 0.010} & \textbf{0.645 $\pm$ 0.015} & 0.903 $\pm$ 0.011          & 0.842 $\pm$ 0.023          & \textbf{0.658 $\pm$ 0.021} & 0.886 $\pm$ 0.006          & \textbf{0.850 $\pm$ 0.004} & \textbf{0.632 $\pm$ 0.014} \\
ModelSoup   & 0.899 $\pm$ 0.006          & \textbf{0.845 $\pm$ 0.009} & 0.638 $\pm$ 0.015          & 0.912 $\pm$ 0.007          & 0.847 $\pm$ 0.016          & \textbf{0.654 $\pm$ 0.018} & \textbf{0.889 $\pm$ 0.010} & \textbf{0.844 $\pm$ 0.011} & 0.622 $\pm$ 0.016          \\
FedAvg      & \textbf{0.904 $\pm$ 0.005} & \textbf{0.855 $\pm$ 0.006} & \textbf{0.642 $\pm$ 0.016} & \textbf{0.922 $\pm$ 0.006} & \textbf{0.866 $\pm$ 0.011} & \textbf{0.662 $\pm$ 0.022} & \textbf{0.888 $\pm$ 0.007} & \textbf{0.845 $\pm$ 0.009} & 0.623 $\pm$ 0.014          \\
\end{tabular}
}

\label{tab:res_patch_het_training_het_test}
\end{table}

\subsubsection{Whole image classifier}

\begin{figure}[h]
	\centering
	\includegraphics[width=0.8\linewidth]{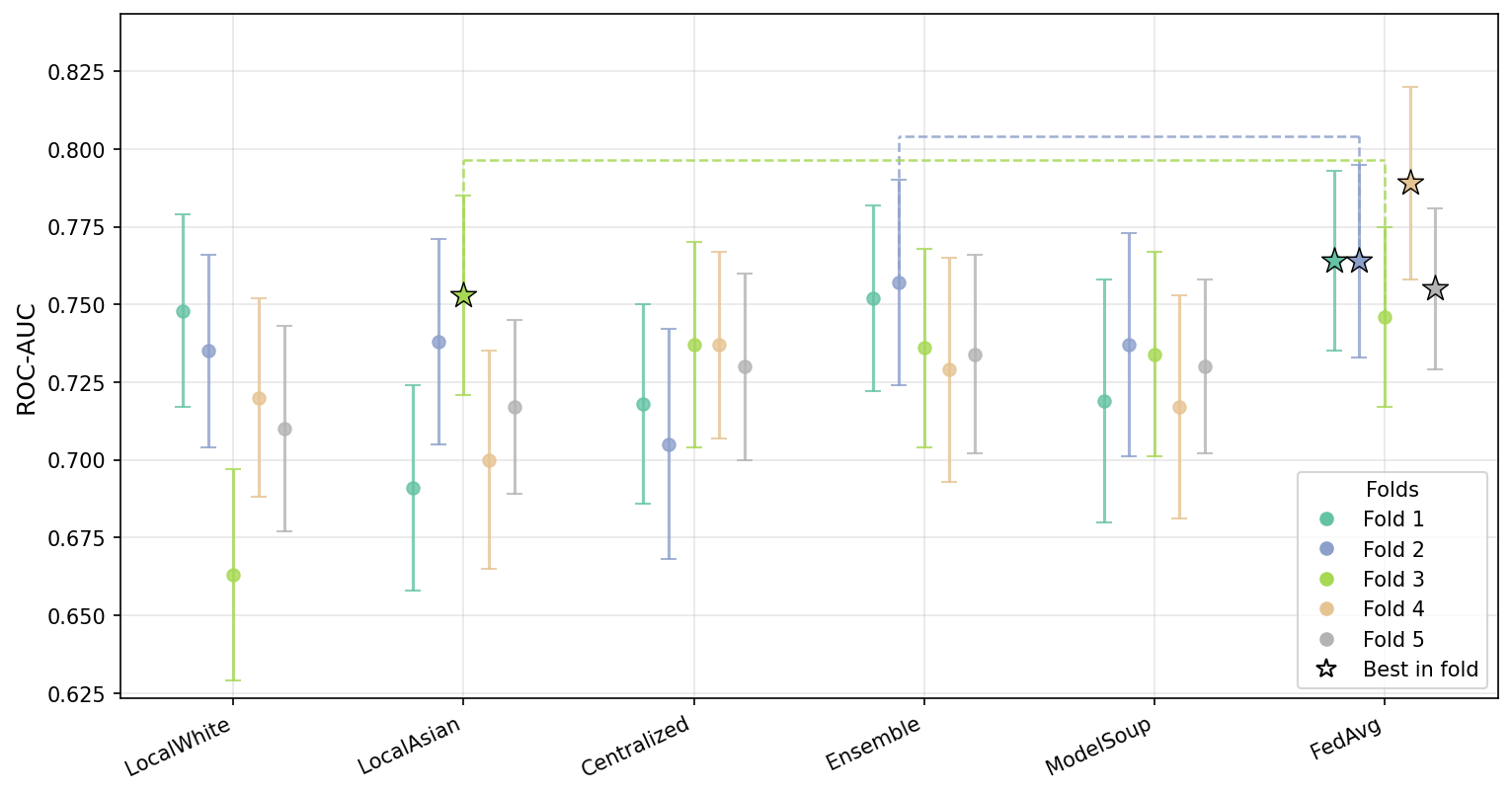}
	\caption{\small Folds bootstrap results of the whole image classifiers trained on the population-based heterogeneous setting and evaluated on the original test set, in terms of the AUC-ROC (mean $\pm$ covariance). Best models for each fold are marked with a star symbol. Models that are not statistically different from the best model according to the Wilcoxon signed-rank test (p-values $>$ 0.1) are indicated by horizontal dashed lines.}
    \label{fig:res_whole_img_het_training_original_test}
\end{figure}

Figure \ref{fig:res_whole_img_het_training_original_test} and Table \ref{tab:realistic_whole_image_roc_auc_std_AsianWhite} show the performance of the whole-image classifiers on the original and population-based test sets, respectively, both by fold and in aggregate. Tables \ref{tab:realitic_roc_auc_std_Asian-SWOR} and \ref{tab:realitic_roc_auc_std_White-SWOR} show the results for the Asian and White population-based test sets, respectively. While Model Ensembling achieves superior performance than centralized training on the original test set, Model Soup outperforms it on the population-based test set, showcasing the positive effects of weight averaging. Finally, the federated model (FedAvg) statistically outperforms centralized training and Model Ensembling and ModelSoup in the two test sets. Additionally, FedAvg outperforms Model Ensembling and ModelSoup in almost all folds. The strong performance of FedAvg can be attributed to the improved weight aggregation mechanism, as elaborated in Section \ref{sec:intro_related_works}. The same trends are observed in the Asian subset of the population-based test set, where FedAvg significantly outperforms in all the folds and in the CV aggregated results. In contrast, within the White population subset, there is no clear best-performing method and no easily identifiable trend across the folds. We attribute this variability to the limited subset size. Further experiments should therefore be conducted on larger White population-based subsets.

% Test set AsianWhite

\begin{table}[H]
    \centering
    \caption{\small Performance of the whole image classifiers trained on the population-based heterogeneous setting with 2 clients and evaluated population-based the test set, in terms of the AUC-ROC (mean $\pm$ covariance). Bold values indicate the best-performing models, whose differences in performance are not statistically significant (Wilcoxon Signed Rank test p-values greater than 0.1).}
    \caption*{\textbf{Population-based setting - Whole Image Clfs. - Population-based AUC-ROC}}
    \resizebox{\textwidth}{!}{
\begin{tabular}{lcccccc}
\toprule
Model       & Fold 1                                   & Fold 2                                   & Fold 3                                   & Fold 4                                   & Fold 5                                   & CV                                  \\
\midrule
LocalWhite  & \textbf{0.741 $\pm$ 0.031 (p $=$ 0.529)} & 0.726 $\pm$ 0.034 (p $<$ 0.010)          & 0.668 $\pm$ 0.038 (p $<$ 0.010)          & 0.728 $\pm$ 0.035 (p $<$ 0.010)          & 0.739 $\pm$ 0.037 (p $<$ 0.010)          & 0.721 $\pm$ 0.030 (p $=$ 0.062)          \\
LocalAsian  & 0.683 $\pm$ 0.043 (p $<$ 0.010)          & 0.731 $\pm$ 0.041 (p $<$ 0.010)          & \textbf{0.770 $\pm$ 0.036 (p $=$ 0.973)} & 0.692 $\pm$ 0.039 (p $<$ 0.010)          & 0.718 $\pm$ 0.040 (p $<$ 0.010)          & \textbf{0.720 $\pm$ 0.033 (p $=$ 0.125)} \\
Centralized & 0.718 $\pm$ 0.040 (p $<$ 0.010)          & 0.725 $\pm$ 0.035 (p $<$ 0.010)          & 0.756 $\pm$ 0.034 (p $<$ 0.010)          & 0.745 $\pm$ 0.035 (p $<$ 0.010)          & \textbf{0.766 $\pm$ 0.036 (p $=$ 1.000)} & 0.739 $\pm$ 0.020 (p $=$ 0.062)          \\
Ensemble    & \textbf{0.740 $\pm$ 0.034 (p $=$ 0.492)} & 0.749 $\pm$ 0.034 (p $<$ 0.010)          & 0.753 $\pm$ 0.031 (p $<$ 0.010)          & 0.736 $\pm$ 0.039 (p $<$ 0.010)          & 0.745 $\pm$ 0.033 (p $<$ 0.010)          & 0.745 $\pm$ 0.008 (p $=$ 0.062)          \\
ModelSoup   & \textbf{0.741 $\pm$ 0.042 (p $=$ 0.550)} & 0.766 $\pm$ 0.038 (p $=$ 0.056)          & \textbf{0.763 $\pm$ 0.030 (p $=$ 0.115)} & 0.763 $\pm$ 0.040 (p $<$ 0.010)          & \textbf{0.761 $\pm$ 0.035 (p $=$ 0.259)} & 0.756 $\pm$ 0.009 (p $=$ 0.062)          \\
FedAvg      & \textbf{0.744 $\pm$ 0.038 (p $=$ 1.000)} & \textbf{0.776 $\pm$ 0.034 (p $=$ 1.000)} & \textbf{0.771 $\pm$ 0.035 (p $=$ 1.000)} & \textbf{0.808 $\pm$ 0.033 (p $=$ 1.000)} & \textbf{0.766 $\pm$ 0.033 (p $=$ 1.000)} & \textbf{0.774 $\pm$ 0.020 (p $=$ 1.000)} \\
\end{tabular}
}

    \label{tab:realistic_whole_image_roc_auc_std_AsianWhite}
\end{table}

% Asian population

\begin{table}[H]
    \centering
    \caption{\small Performance of the whole image classifiers trained on the population-based heterogeneous setting with 2 clients and evaluated the Asian population-based test set, in terms of the AUC-ROC (mean $\pm$ covariance). Bold values indicate the best-performing models, whose differences in performance are not statistically significant (Wilcoxon Signed Rank test p-values greater than 0.1).}
    \caption*{\textbf{Population-based setting - Whole Image Clfs. - Asian population-based AUC-ROC}}
    \resizebox{\textwidth}{!}{
\begin{tabular}{lcccccc}
\toprule
Model       & Fold 1                                   & Fold 2                                   & Fold 3                                   & Fold 4                                   & Fold 5                                   & CV                                  \\
\midrule
LocalWhite  & 0.703 $\pm$ 0.051 (p $<$ 0.010)          & 0.719 $\pm$ 0.044 (p $<$ 0.010)          & 0.673 $\pm$ 0.056 (p $<$ 0.010)          & 0.734 $\pm$ 0.048 (p $<$ 0.010)          & 0.685 $\pm$ 0.053 (p $<$ 0.010)          & 0.703 $\pm$ 0.025 (p $=$ 0.062)          \\
LocalAsian  & 0.675 $\pm$ 0.055 (p $<$ 0.010)          & 0.735 $\pm$ 0.048 (p $<$ 0.010)          & 0.726 $\pm$ 0.051 (p $<$ 0.010)          & 0.661 $\pm$ 0.072 (p $<$ 0.010)          & 0.699 $\pm$ 0.049 (p $<$ 0.010)          & 0.697 $\pm$ 0.036 (p $=$ 0.062)          \\
Centralized & 0.733 $\pm$ 0.054 (p $<$ 0.010)          & 0.738 $\pm$ 0.061 (p $<$ 0.010)          & 0.678 $\pm$ 0.068 (p $<$ 0.010)          & 0.752 $\pm$ 0.055 (p $<$ 0.010)          & 0.712 $\pm$ 0.045 (p $<$ 0.010)          & 0.716 $\pm$ 0.029 (p $=$ 0.062)          \\
Ensemble    & 0.689 $\pm$ 0.057 (p $<$ 0.010)          & 0.730 $\pm$ 0.053 (p $<$ 0.010)          & 0.721 $\pm$ 0.049 (p $<$ 0.010)          & 0.715 $\pm$ 0.055 (p $<$ 0.010)          & 0.693 $\pm$ 0.058 (p $<$ 0.010)          & 0.713 $\pm$ 0.019 (p $=$ 0.062)          \\
ModelSoup   & 0.724 $\pm$ 0.058 (p $<$ 0.010)          & 0.713 $\pm$ 0.054 (p $<$ 0.010)          & 0.722 $\pm$ 0.047 (p $<$ 0.010)          & 0.732 $\pm$ 0.054 (p $<$ 0.010)          & 0.733 $\pm$ 0.046 (p $<$ 0.010)          & 0.730 $\pm$ 0.006 (p $=$ 0.062)          \\
FedAvg      & \textbf{0.767 $\pm$ 0.048 (p $=$ 1.000)} & \textbf{0.765 $\pm$ 0.054 (p $=$ 1.000)} & \textbf{0.783 $\pm$ 0.043 (p $=$ 1.000)} & \textbf{0.797 $\pm$ 0.043 (p $=$ 1.000)} & \textbf{0.763 $\pm$ 0.037 (p $=$ 1.000)} & \textbf{0.774 $\pm$ 0.017 (p $=$ 1.000)} \\
\end{tabular}
}

    \label{tab:realitic_roc_auc_std_Asian-SWOR}
\end{table}

% White population

\begin{table}[h]
    \centering
    \caption{\small Performance of the whole image classifiers trained on the population-based heterogeneous setting with 2 clients and evaluated on the White population-based test set, in terms of the AUC-ROC (mean $\pm$ covariance). Bold values indicate the best-performing models, whose differences in performance are not statistically significant (Wilcoxon Signed Rank test p-values greater than 0.1).}
    \caption*{\textbf{Population-based setting - Whole Image Clfs. - White population-based AUC-ROC}}
    \resizebox{\textwidth}{!}{
\begin{tabular}{lcccccc}
\toprule
Model       & Fold 1                                   & Fold 2                                   & Fold 3                                   & Fold 4                                   & Fold 5                                   & CV                                  \\
\midrule
LocalWhite  & \textbf{0.802 $\pm$ 0.040 (p $=$ 1.000)} & 0.744 $\pm$ 0.045 (p $<$ 0.010)          & 0.665 $\pm$ 0.058 (p $<$ 0.010)          & 0.732 $\pm$ 0.055 (p $<$ 0.010)          & 0.800 $\pm$ 0.041 (p $=$ 0.024)          & \textbf{0.746 $\pm$ 0.056 (p $=$ 0.312)} \\
LocalAsian  & 0.725 $\pm$ 0.055 (p $<$ 0.010)          & 0.740 $\pm$ 0.056 (p $<$ 0.010)          & 0.813 $\pm$ 0.041 (p $<$ 0.010)          & 0.746 $\pm$ 0.048 (p $<$ 0.010)          & 0.729 $\pm$ 0.051 (p $<$ 0.010)          & \textbf{0.749 $\pm$ 0.038 (p $=$ 0.125)} \\
Centralized & 0.703 $\pm$ 0.054 (p $<$ 0.010)          & 0.715 $\pm$ 0.053 (p $<$ 0.010)          & \textbf{0.850 $\pm$ 0.036 (p $=$ 1.000)} & 0.760 $\pm$ 0.048 (p $<$ 0.010)          & \textbf{0.811 $\pm$ 0.048 (p $=$ 1.000)} & \textbf{0.764 $\pm$ 0.062 (p $=$ 0.438)} \\
Ensemble    & \textbf{0.796 $\pm$ 0.047 (p $=$ 0.417)} & 0.786 $\pm$ 0.049 (p $<$ 0.010)          & 0.787 $\pm$ 0.041 (p $<$ 0.010)          & 0.763 $\pm$ 0.049 (p $<$ 0.010)          & 0.790 $\pm$ 0.042 (p $<$ 0.010)          & \textbf{0.784 $\pm$ 0.016 (p $=$ 0.812)} \\
ModelSoup   & 0.758 $\pm$ 0.059 (p $<$ 0.010)          & \textbf{0.821 $\pm$ 0.044 (p $=$ 1.000)} & 0.806 $\pm$ 0.043 (p $<$ 0.010)          & 0.789 $\pm$ 0.055 (p $<$ 0.010)          & 0.794 $\pm$ 0.045 (p $<$ 0.010)          & \textbf{0.788 $\pm$ 0.023 (p $=$ 1.000)} \\
FedAvg      & 0.746 $\pm$ 0.047 (p $<$ 0.010)          & 0.779 $\pm$ 0.054 (p $<$ 0.010)          & 0.744 $\pm$ 0.055 (p $<$ 0.010)          & \textbf{0.827 $\pm$ 0.041 (p $=$ 1.000)} & 0.789 $\pm$ 0.048 (p $<$ 0.010)          & \textbf{0.780 $\pm$ 0.031 (p $=$ 0.438)} \\
\end{tabular}
}

    \label{tab:realitic_roc_auc_std_White-SWOR}
\end{table}

\subsection{Strongly heterogeneous setting - 2 clients}

We now present and discuss the results of training in the strongly heterogeneous setting, with evaluation on the original test set.

\subsubsection{Patch-classifier}

Table \ref{tab:res_patch_strongly_het} presents the results on the patch-classifiers. In addition to FedAvg, two other FL algorithms that tackle data heterogeneity are considered: FedProx and SCAFFOLD. As expected, the models trained locally (both in the low density and high density sites) underperform with respect to the centralized setting. Besides, the Ensemble and Model Soup models are unable to match the performance of the centralized setting, which shows the inefficacy of simple Ensembling and weight averaging approaches when heterogeneity is strong. Finally, all the FL approaches achieve performance comparable to the centralized setting, with no statistically significant difference. This suggests that the data heterogeneity resulting from breast density differences does not necessitate the use of FL algorithms specifically designed to handle heterogeneity, even in highly heterogeneous scenarios. This also justifies the decision to exclude such algorithms from the experiments in Section~\ref{sec:res_pop_based_het}. These conclusions hold when analyzing the results on the low- and high-density subsets of the test set (see the two rightmost columns in Table~\ref{tab:res_patch_strongly_het}). Section \ref{app:sec:strong_het_2_clients_patch} in the Appendix contains the per-fold performance for each of the CV aggregated performances in Table \ref{tab:res_patch_strongly_het}.

\begin{table}[H]
\centering
\caption{\small Performance of the patch-classifiers trained on the strongly heterogeneous setting and evaluated on the density-based test sets (mean $\pm$ covariance). Bolded values indicate the best-performing models, whose differences in performance are not statistically significant (Wilcoxon Signed Rank test p-values greater than 0.1)}
\caption*{\textbf{Strong setting (2 clients) - Patch Clfs. - Performance}}
\resizebox{\textwidth}{!}{
\begin{tabular}{lccccccccc}
\toprule
Model       & \multicolumn{3}{c}{Overall}                                                          & \multicolumn{3}{c}{Low densities population}                                                              & \multicolumn{3}{c}{High densities population}                                                             \\
\cmidrule(lr){2-4} \cmidrule(lr){5-7} \cmidrule(lr){8-10}
            & AUC OvO                    & AUC OvR                    & Accuracy                   & AUC OvO                    & AUC OvR                    & Accuracy                   & AUC OvO                    & AUC OvR                    & Accuracy                   \\
\midrule
LocalLow    & 0.901 $\pm$ 0.007          & 0.849 $\pm$ 0.011          & 0.638 $\pm$ 0.019          & 0.891 $\pm$ 0.009          & 0.844 $\pm$ 0.016          & 0.669 $\pm$ 0.014          & 0.901 $\pm$ 0.006          & 0.847 $\pm$ 0.010          & 0.623 $\pm$ 0.023          \\
LocalHigh   & 0.902 $\pm$ 0.001          & 0.855 $\pm$ 0.005          & 0.627 $\pm$ 0.019          & 0.877 $\pm$ 0.006          & 0.839 $\pm$ 0.013          & 0.632 $\pm$ 0.024          & 0.906 $\pm$ 0.002          & 0.857 $\pm$ 0.003          & 0.625 $\pm$ 0.023          \\
Centralized & \textbf{0.918 $\pm$ 0.004} & \textbf{0.874 $\pm$ 0.003} & \textbf{0.676 $\pm$ 0.007} & \textbf{0.905 $\pm$ 0.006} & \textbf{0.864 $\pm$ 0.010} & \textbf{0.695 $\pm$ 0.015} & \textbf{0.919 $\pm$ 0.004} & \textbf{0.874 $\pm$ 0.004} & \textbf{0.666 $\pm$ 0.008} \\
Ensemble    & 0.898 $\pm$ 0.003          & 0.856 $\pm$ 0.004          & 0.651 $\pm$ 0.014          & 0.873 $\pm$ 0.006          & 0.840 $\pm$ 0.010          & 0.660 $\pm$ 0.016          & 0.902 $\pm$ 0.003          & 0.858 $\pm$ 0.003          & 0.647 $\pm$ 0.014          \\
ModelSoup   & \textbf{0.907 $\pm$ 0.005} & 0.857 $\pm$ 0.007          & 0.637 $\pm$ 0.015          & \textbf{0.887 $\pm$ 0.009} & 0.845 $\pm$ 0.016          & 0.644 $\pm$ 0.016          & 0.909 $\pm$ 0.003          & 0.859 $\pm$ 0.003          & 0.633 $\pm$ 0.014          \\
FedAvg      & 0.915 $\pm$ 0.004          & \textbf{0.874 $\pm$ 0.004} & \textbf{0.671 $\pm$ 0.004} & \textbf{0.902 $\pm$ 0.008} & \textbf{0.861 $\pm$ 0.014} & \textbf{0.690 $\pm$ 0.013} & 0.916 $\pm$ 0.003          & \textbf{0.877 $\pm$ 0.002} & \textbf{0.661 $\pm$ 0.007} \\
FedProx     & 0.911 $\pm$ 0.005          & \textbf{0.875 $\pm$ 0.004} & \textbf{0.670 $\pm$ 0.010} & 0.898 $\pm$ 0.007          & \textbf{0.864 $\pm$ 0.007} & \textbf{0.689 $\pm$ 0.018} & 0.912 $\pm$ 0.005          & \textbf{0.877 $\pm$ 0.003} & \textbf{0.660 $\pm$ 0.008} \\
SCAFFOLD    & \textbf{0.914 $\pm$ 0.002} & \textbf{0.875 $\pm$ 0.003} & 0.666 $\pm$ 0.007          & \textbf{0.900 $\pm$ 0.005} & \textbf{0.863 $\pm$ 0.011} & \textbf{0.686 $\pm$ 0.013} & \textbf{0.915 $\pm$ 0.003} & \textbf{0.876 $\pm$ 0.005} & 0.656 $\pm$ 0.009          \\
\end{tabular}
}

\label{tab:res_patch_strongly_het}
\end{table}

\subsubsection{Whole Image classifier}
\label{sec:strong_het_whole_img_results}

Table \ref{tab:strong_2_clients_roc_auc_std_Full} shows the per-fold bootstrap and CV results of the whole image classifiers. When considering the CV aggregated performance, we can see that the Ensembling, Model Soup, and federated models match the performance of the centralized setting, outperforming the local trainings. When comparing local trainings, we can see a big difference between the low and high density sites. We hypothesize that the decreased visibility of lesions in high density images negatively impacts training, especially when they are not combined with low density images, making it harder for the model to learn semantic concepts from the data. In contrast, training on low-density images allows the model to generalize effectively, even when applied to high-density data. 

When analyzing the per-fold results, we can see that Ensemble and FedAvg outperform the low density site in every fold, which shows the limitations of using statistical tests for 5-fold CV results. Overall, the low density site is outperformed by all the non-local models in folds 1, 3, 4 and 5, and manages to significantly outperform Centralized, ModelSoup and FedProx in fold 2. The high density site is outperformed by the non-local models in all the folds. The fold-wise performance of the non-local models aligns with their p-values, suggesting no meaningful performance differences between them.

% Test set Full

\begin{table}[H]
    \centering
    \caption{\small Performance of the whole image classifiers trained on the strongly heterogeneous setting with 2 clients and evaluated on the originally test set, in terms of the AUC-ROC  (mean $\pm$ covariance). Bold values indicate the best-performing models, whose differences in performance are not statistically significant (Wilcoxon Signed Rank test p-values greater than 0.1).}
    \caption*{\textbf{Strong setting (2 clients) - Whole Image Clfs. - Overall AUC-ROC}}
    \resizebox{\textwidth}{!}{
\begin{tabular}{lcccccc}
\toprule
Model       & Fold 1                                   & Fold 2                                   & Fold 3                                   & Fold 4                                   & Fold 5                                   & CV                                  \\
\midrule
LocalLow    & 0.721 $\pm$ 0.028 (p $<$ 0.010)          & 0.768 $\pm$ 0.024 (p $<$ 0.010)          & 0.755 $\pm$ 0.028 (p $<$ 0.010)          & 0.751 $\pm$ 0.031 (p $=$ 0.013)          & 0.733 $\pm$ 0.029 (p $<$ 0.010)          & 0.743 $\pm$ 0.020 (p $=$ 0.062)          \\
LocalHigh   & 0.715 $\pm$ 0.036 (p $<$ 0.010)          & 0.724 $\pm$ 0.030 (p $<$ 0.010)          & 0.744 $\pm$ 0.030 (p $<$ 0.010)          & 0.721 $\pm$ 0.035 (p $<$ 0.010)          & 0.744 $\pm$ 0.032 (p $<$ 0.010)          & 0.731 $\pm$ 0.014 (p $=$ 0.062)          \\
Centralized & 0.758 $\pm$ 0.026 (p $<$ 0.010)          & 0.725 $\pm$ 0.029 (p $<$ 0.010)          & 0.756 $\pm$ 0.028 (p $<$ 0.010)          & \textbf{0.756 $\pm$ 0.030 (p $=$ 0.332)} & 0.762 $\pm$ 0.028 (p $<$ 0.010)          & \textbf{0.751 $\pm$ 0.013 (p $=$ 0.312)} \\
Ensemble    & 0.745 $\pm$ 0.027 (p $<$ 0.010)          & \textbf{0.788 $\pm$ 0.024 (p $=$ 1.000)} & 0.777 $\pm$ 0.028 (p $=$ 0.089)          & \textbf{0.762 $\pm$ 0.027 (p $=$ 1.000)} & 0.765 $\pm$ 0.028 (p $<$ 0.010)          & \textbf{0.766 $\pm$ 0.017 (p $=$ 1.000)} \\
ModelSoup   & 0.690 $\pm$ 0.032 (p $<$ 0.010)          & 0.760 $\pm$ 0.026 (p $<$ 0.010)          & \textbf{0.785 $\pm$ 0.027 (p $=$ 1.000)} & 0.739 $\pm$ 0.027 (p $<$ 0.010)          & 0.771 $\pm$ 0.028 (p $=$ 0.042)          & \textbf{0.748 $\pm$ 0.038 (p $=$ 0.312)} \\
FedAvg      & \textbf{0.779 $\pm$ 0.026 (p $=$ 1.000)} & 0.764 $\pm$ 0.028 (p $<$ 0.010)          & 0.774 $\pm$ 0.026 (p $<$ 0.010)          & 0.750 $\pm$ 0.033 (p $=$ 0.031)          & 0.763 $\pm$ 0.032 (p $<$ 0.010)          & \textbf{0.766 $\pm$ 0.012 (p $=$ 1.000)} \\
FedProx     & 0.762 $\pm$ 0.029 (p $<$ 0.010)          & 0.729 $\pm$ 0.036 (p $<$ 0.010)          & 0.768 $\pm$ 0.030 (p $<$ 0.010)          & \textbf{0.758 $\pm$ 0.037 (p $=$ 0.343)} & 0.750 $\pm$ 0.033 (p $<$ 0.010)          & \textbf{0.753 $\pm$ 0.014 (p $=$ 0.438)} \\
SCAFFOLD    & 0.770 $\pm$ 0.030 (p $=$ 0.013)          & 0.757 $\pm$ 0.024 (p $<$ 0.010)          & 0.762 $\pm$ 0.034 (p $<$ 0.010)          & 0.746 $\pm$ 0.029 (p $<$ 0.010)          & \textbf{0.781 $\pm$ 0.036 (p $=$ 1.000)} & \textbf{0.765 $\pm$ 0.016 (p $=$ 1.000)} \\
\end{tabular}
}

    \label{tab:strong_2_clients_roc_auc_std_Full}
\end{table}

\subsection{Strongly heterogeneous setting - 4 clients}

\subsubsection{Whole image classifier}

Table \ref{tab:strong_4_clients_roc_auc_std_Full} compares the models in terms of their per-fold performance and in their aggregated Cross Validation performance. Federated models (FedAvg, FedProx, and SCAFFOLD) significantly outperform the local models, and they also tend to outperform the centralized and ensemble models, although these differences are not statistically significant (Wilcoxon Signed Rank test p-values greater than 0.1). This trend is consistent at the fold level: federated models outperform local models in all folds and surpass the centralized model in 4 out of 5 folds. Notably, under this extreme heterogeneity setting, the Model Soup approach fails entirely, yielding the lowest performance among all evaluated models. As this setting is more heterogeneous than the previous ones, naively averaging the weights of models trained on different subsets causes performance to drop sharply. Of the local models, the one trained on Density B images shows superior performance. This may be explained by the relative ease of learning from lower-density images, where lesions are more visible, and is consistent with what was observed in Section \ref{sec:strong_het_whole_img_results}. Finally, FedAvg remains robust under this highly heterogeneous setting, outperforming all other models in aggregated cross-validation performance and achieving the best results in 3 of the 5 folds.

\begin{table}[H]
\centering
\caption{\small Performance of the whole image classifiers trained on the strongly heterogeneous setting with 4 clients and evaluated on the original test set, in terms of the AUC-ROC (mean $\pm$ covariance). Bold values indicate the best-performing models, whose differences in performance are not statistically significant (Wilcoxon Signed Rank test p-values greater than 0.1).}
\caption*{\textbf{Strong setting (4 clients) - Whole Image Clfs. - Overall AUC-ROC}}
\resizebox{\textwidth}{!}{
\begin{tabular}{lcccccc}
\toprule
Model       & Fold 1                                   & Fold 2                                   & Fold 3                                   & Fold 4                                   & Fold 5                                   & CV                                  \\
\midrule
A           & 0.652 $\pm$ 0.036 (p $<$ 0.010)          & 0.525 $\pm$ 0.035 (p $<$ 0.010)          & 0.530 $\pm$ 0.032 (p $<$ 0.010)          & 0.686 $\pm$ 0.034 (p $<$ 0.010)          & 0.583 $\pm$ 0.032 (p $<$ 0.010)          & 0.593 $\pm$ 0.071 (p $=$ 0.062)          \\
B           & 0.729 $\pm$ 0.027 (p $<$ 0.010)          & 0.755 $\pm$ 0.027 (p $<$ 0.010)          & 0.719 $\pm$ 0.030 (p $<$ 0.010)          & 0.702 $\pm$ 0.030 (p $<$ 0.010)          & 0.739 $\pm$ 0.026 (p $<$ 0.010)          & 0.730 $\pm$ 0.021 (p $=$ 0.062)          \\
C           & 0.718 $\pm$ 0.031 (p $<$ 0.010)          & 0.699 $\pm$ 0.032 (p $<$ 0.010)          & 0.672 $\pm$ 0.032 (p $<$ 0.010)          & 0.701 $\pm$ 0.038 (p $<$ 0.010)          & 0.713 $\pm$ 0.030 (p $<$ 0.010)          & 0.697 $\pm$ 0.016 (p $=$ 0.062)          \\
D           & 0.720 $\pm$ 0.037 (p $<$ 0.010)          & 0.667 $\pm$ 0.033 (p $<$ 0.010)          & 0.705 $\pm$ 0.036 (p $<$ 0.010)          & 0.691 $\pm$ 0.031 (p $<$ 0.010)          & 0.704 $\pm$ 0.032 (p $<$ 0.010)          & 0.696 $\pm$ 0.018 (p $=$ 0.062)          \\
Centralized & 0.759 $\pm$ 0.029 (p $<$ 0.010)          & 0.728 $\pm$ 0.028 (p $<$ 0.010)          & 0.760 $\pm$ 0.028 (p $=$ 0.078)          & \textbf{0.754 $\pm$ 0.029 (p $=$ 1.000)} & 0.761 $\pm$ 0.027 (p $<$ 0.010)          & \textbf{0.751 $\pm$ 0.013 (p $=$ 0.438)} \\
Ensemble    & 0.762 $\pm$ 0.028 (p $<$ 0.010)          & 0.755 $\pm$ 0.030 (p $=$ 0.038)          & 0.730 $\pm$ 0.029 (p $<$ 0.010)          & 0.742 $\pm$ 0.029 (p $<$ 0.010)          & 0.762 $\pm$ 0.028 (p $<$ 0.010)          & \textbf{0.750 $\pm$ 0.012 (p $=$ 0.312)} \\
ModelSoup   & 0.535 $\pm$ 0.040 (p $<$ 0.010)          & 0.641 $\pm$ 0.033 (p $<$ 0.010)          & 0.401 $\pm$ 0.033 (p $<$ 0.010)          & 0.641 $\pm$ 0.031 (p $<$ 0.010)          & 0.440 $\pm$ 0.035 (p $<$ 0.010)          & 0.531 $\pm$ 0.111 (p $=$ 0.062)          \\
FedAvg      & 0.763 $\pm$ 0.030 (p $<$ 0.010)          & \textbf{0.765 $\pm$ 0.023 (p $=$ 1.000)} & \textbf{0.768 $\pm$ 0.030 (p $=$ 1.000)} & 0.721 $\pm$ 0.031 (p $<$ 0.010)          & \textbf{0.791 $\pm$ 0.031 (p $=$ 1.000)} & \textbf{0.763 $\pm$ 0.027 (p $=$ 1.000)} \\
FedProx     & 0.758 $\pm$ 0.032 (p $<$ 0.010)          & 0.738 $\pm$ 0.030 (p $<$ 0.010)          & \textbf{0.764 $\pm$ 0.031 (p $=$ 0.481)} & 0.743 $\pm$ 0.031 (p $=$ 0.059)          & 0.773 $\pm$ 0.029 (p $<$ 0.010)          & \textbf{0.756 $\pm$ 0.015 (p $=$ 0.438)} \\
SCAFFOLD    & \textbf{0.783 $\pm$ 0.034 (p $=$ 1.000)} & 0.747 $\pm$ 0.028 (p $<$ 0.010)          & 0.757 $\pm$ 0.031 (p $=$ 0.017)          & 0.744 $\pm$ 0.034 (p $=$ 0.028)          & 0.768 $\pm$ 0.031 (p $<$ 0.010)          & \textbf{0.761 $\pm$ 0.015 (p $=$ 1.000)} \\
\end{tabular}
}

\label{tab:strong_4_clients_roc_auc_std_Full}
\end{table}

\section{Conclusions and discussion}

In this work, we investigated the impact of breast density–induced data heterogeneity on Federated Learning (FL) for mammography image classification. We explored two realistic heterogeneous scenarios: (1) a strongly heterogeneous setting, where each site exclusively contained images from either low- or high-density patients; and (2) a population-based heterogeneous setting, where sites reflected the breast density distribution observed in White and Asian populations. For the strongly heterogeneous setting, we considered configurations with 2 and 4 clients. FL models were evaluated against centralized training, local training, model Ensembling, and weight averaging (Model Soup).

\begin{sloppypar}
In the strongly heterogeneous setting, local models trained on low- or high-density data exhibited significantly lower performance than centralized models, underscoring the limitations of isolated learning under restricted data diversity. Simple aggregation methods such as Ensembling and Model Soup did not recover centralized performance in patch-level classification tasks, suggesting their limitations in handling severe distribution shifts. However, for whole image classification, Ensembling and Model Soup approached or slightly exceeded centralized performance in the 2-client setting, although differences were not always statistically significant. In the 4-client setting, while Ensembling achieved competitive performance, Model Soup performed poorly, highlighting the limitations of naively averaging the weights of models trained independently on heterogeneous subsets of the data. Standard FL approaches—FedAvg, FedProx, and SCAFFOLD—consistently matched, and in some cases exceeded, centralized performance across both patch-level and image-level tasks. Notably, we did not observe statistically significant benefits from heterogeneity-aware methods (FedProx, SCAFFOLD) over the baseline FedAvg, suggesting that breast density heterogeneity, while nontrivial, may not be severe enough in this context to warrant more complex FL strategies. A key advantage of these federated methods is that they maintain the same number of neural network evaluations as centralized learning, unlike Ensembling, where evaluations increase with the number of training subsets.
\end{sloppypar}

\begin{sloppypar}
In the population-based heterogeneous setting, FedAvg was the only method to consistently match or exceed centralized performance across both test sets and tasks. Remarkably, FedAvg outperformed centralized training for whole image classification. We hypothesize that repeated rounds of local updates and aggregation in FL may act as a refined form of weight averaging, enhancing generalization in heterogeneous settings—a hypothesis warranting further empirical and theoretical validation. In contrast, Ensembling and Model Soup failed to consistently match centralized performance, despite occasional strong results. Local training on the White population resulted in poor generalization, while training on the Asian population yielded relatively strong results on the population-based patch-level test set. However, its performance dropped on the original test set and for whole image classification, indicating limited generalizability.
\end{sloppypar}

Overall, our findings support the applicability of standard FL methods, particularly FedAvg, in mammography classification tasks under population-based heterogeneous and strongly heterogeneous conditions driven by breast density. These results underscore the promise of FL for real-world deployment in collaborative medical imaging applications. An important limitation of this study is the use of 5 folds, selected due to resource constraints, which reduces the statistical power of the analyses and limits the use of statistical tests. In future work, we plan to increase the number of folds to confirm and extend the current findings. Another limitation of this work is that, in the population-based heterogeneity setting, the split was based exclusively on density information, since ethnicity information was unavailable in our dataset. Investigating population heterogeneity through explicit ethnicity-based modeling and experimental validation is left for future work. Future work should also explore the generalizability of these conclusions to other tasks such as lesion detection and segmentation, to larger and more diverse federated networks, and to privacy-preserving FL methods such as Differential Privacy, Secure Aggregation, and Homomorphic Encryption \cite{kairouz_FL_review}. Finally, future studies could investigate the combined impact of multiple sources of heterogeneity in a Federated Learning setting, including breast density, dataset size, and cancer or lesion distribution, to better understand their interactions and influence on model performance.

\subsection{Appendix}

\subsubsection{Population-based heterogeneous setting - 2 clients}

\paragraph{Patch classifier}

% Overall test set

\begin{table}[H]
    \centering
    \caption{\small Performance of the patch classifiers trained on the population-based heterogeneous setting with 2 clients and evaluated on the original test set, in terms of the AUC-ROC One-vs-One (mean $\pm$ covariance). Bold values indicate the best-performing models, whose differences in performance are not statistically significant (Wilcoxon Signed Rank test p-values greater than 0.1).}
    \caption*{\textbf{Population-based setting - Patch Clfs. - Overall AUC-ROC OvO}}
    \resizebox{\textwidth}{!}{
\begin{tabular}{lcccccc}
\toprule
Model       & Fold 1                                   & Fold 2                                   & Fold 3                                   & Fold 4                                   & Fold 5                                   & CV                                  \\
\midrule
LocalWhite  & 0.875 $\pm$ 0.005 (p $<$ 0.010)          & 0.880 $\pm$ 0.005 (p $<$ 0.010)          & 0.891 $\pm$ 0.004 (p $<$ 0.010)          & 0.896 $\pm$ 0.004 (p $<$ 0.010)          & 0.884 $\pm$ 0.004 (p $<$ 0.010)          & 0.885 $\pm$ 0.009 (p $=$ 0.062)          \\
LocalAsian  & 0.897 $\pm$ 0.004 (p $<$ 0.010)          & 0.903 $\pm$ 0.005 (p $<$ 0.010)          & 0.904 $\pm$ 0.004 (p $<$ 0.010)          & 0.906 $\pm$ 0.004 (p $<$ 0.010)          & 0.900 $\pm$ 0.004 (p $<$ 0.010)          & 0.902 $\pm$ 0.003 (p $=$ 0.062)          \\
Centralized & \textbf{0.900 $\pm$ 0.005 (p $=$ 1.000)} & \textbf{0.909 $\pm$ 0.004 (p $=$ 1.000)} & \textbf{0.913 $\pm$ 0.004 (p $=$ 1.000)} & \textbf{0.910 $\pm$ 0.004 (p $=$ 1.000)} & \textbf{0.903 $\pm$ 0.004 (p $=$ 1.000)} & \textbf{0.906 $\pm$ 0.005 (p $=$ 1.000)} \\
Ensemble    & 0.886 $\pm$ 0.004 (p $<$ 0.010)          & 0.896 $\pm$ 0.005 (p $<$ 0.010)          & 0.895 $\pm$ 0.004 (p $<$ 0.010)          & 0.904 $\pm$ 0.004 (p $<$ 0.010)          & 0.890 $\pm$ 0.005 (p $<$ 0.010)          & 0.894 $\pm$ 0.007 (p $=$ 0.062)          \\
ModelSoup   & 0.892 $\pm$ 0.004 (p $<$ 0.010)          & 0.902 $\pm$ 0.004 (p $<$ 0.010)          & 0.905 $\pm$ 0.004 (p $<$ 0.010)          & 0.904 $\pm$ 0.004 (p $<$ 0.010)          & 0.896 $\pm$ 0.004 (p $<$ 0.010)          & 0.900 $\pm$ 0.006 (p $=$ 0.062)          \\
FedAvg      & \textbf{0.899 $\pm$ 0.004 (p $=$ 0.147)} & 0.907 $\pm$ 0.004 (p $<$ 0.010)          & 0.911 $\pm$ 0.004 (p $<$ 0.010)          & 0.908 $\pm$ 0.004 (p $<$ 0.010)          & \textbf{0.903 $\pm$ 0.004 (p $=$ 1.000)} & \textbf{0.905 $\pm$ 0.005 (p $=$ 0.125)} \\
\end{tabular}
}

    \label{tab:realistic_patch_roc_auc_ovo_std_Full}
\end{table}

\begin{table}[H]
    \centering
    \caption{\small Performance of the patch classifiers trained on the population-based heterogeneous setting with 2 clients and evaluated on the original test set, in terms of the AUC-ROC One-vs-Rest (mean $\pm$ covariance). Bold values indicate the best-performing models, whose differences in performance are not statistically significant (Wilcoxon Signed Rank test p-values greater than 0.1).}
    \caption*{\textbf{Population-based setting - Patch Clfs. - Overall AUC-ROC OvR}}
    \resizebox{\textwidth}{!}{
\begin{tabular}{lcccccc}
\toprule
Model       & Fold 1                                   & Fold 2                                   & Fold 3                                   & Fold 4                                   & Fold 5                                   & CV                                  \\
\midrule
LocalWhite  & 0.811 $\pm$ 0.006 (p $<$ 0.010)          & 0.828 $\pm$ 0.006 (p $<$ 0.010)          & 0.833 $\pm$ 0.006 (p $<$ 0.010)          & 0.840 $\pm$ 0.005 (p $<$ 0.010)          & 0.819 $\pm$ 0.006 (p $<$ 0.010)          & 0.826 $\pm$ 0.012 (p $=$ 0.062)          \\
LocalAsian  & 0.844 $\pm$ 0.006 (p $<$ 0.010)          & 0.854 $\pm$ 0.006 (p $<$ 0.010)          & 0.842 $\pm$ 0.006 (p $<$ 0.010)          & 0.859 $\pm$ 0.005 (p $=$ 0.030)          & 0.849 $\pm$ 0.005 (p $<$ 0.010)          & 0.849 $\pm$ 0.007 (p $=$ 0.062)          \\
Centralized & \textbf{0.852 $\pm$ 0.006 (p $=$ 1.000)} & 0.864 $\pm$ 0.005 (p $=$ 0.015)          & 0.854 $\pm$ 0.006 (p $<$ 0.010)          & 0.853 $\pm$ 0.006 (p $<$ 0.010)          & 0.846 $\pm$ 0.006 (p $<$ 0.010)          & \textbf{0.853 $\pm$ 0.007 (p $=$ 0.125)} \\
Ensemble    & 0.849 $\pm$ 0.006 (p $<$ 0.010)          & 0.860 $\pm$ 0.006 (p $<$ 0.010)          & 0.838 $\pm$ 0.006 (p $<$ 0.010)          & \textbf{0.861 $\pm$ 0.005 (p $=$ 1.000)} & 0.845 $\pm$ 0.006 (p $<$ 0.010)          & \textbf{0.851 $\pm$ 0.010 (p $=$ 0.125)} \\
ModelSoup   & 0.843 $\pm$ 0.006 (p $<$ 0.010)          & 0.856 $\pm$ 0.005 (p $<$ 0.010)          & 0.848 $\pm$ 0.006 (p $<$ 0.010)          & 0.858 $\pm$ 0.005 (p $<$ 0.010)          & 0.842 $\pm$ 0.005 (p $<$ 0.010)          & 0.850 $\pm$ 0.008 (p $=$ 0.062)          \\
FedAvg      & \textbf{0.852 $\pm$ 0.005 (p $=$ 1.000)} & \textbf{0.866 $\pm$ 0.005 (p $=$ 1.000)} & \textbf{0.860 $\pm$ 0.006 (p $=$ 1.000)} & \textbf{0.860 $\pm$ 0.005 (p $=$ 0.205)} & \textbf{0.857 $\pm$ 0.005 (p $=$ 1.000)} & \textbf{0.859 $\pm$ 0.005 (p $=$ 1.000)} \\
\end{tabular}
}

    \label{tab:realistic_patch_roc_auc_AUC-ROC OvR_std_Full}
\end{table}

\begin{table}[H]
    \centering
    \caption{\small Performance of the patch classifiers trained on the population-based heterogeneous setting with 2 clients and evaluated on the original test set, in terms of the accuracy (mean $\pm$ covariance). Bold values indicate the best-performing models, whose differences in performance are not statistically significant (Wilcoxon Signed Rank test p-values greater than 0.1).}
    \caption*{\textbf{Population-based setting - Patch Clfs. - Overall Accuracy}}
    \resizebox{\textwidth}{!}{
\begin{tabular}{lcccccc}
\toprule
Model       & Fold 1                                   & Fold 2                                   & Fold 3                                   & Fold 4                                   & Fold 5                                   & CV                                  \\
\midrule
LocalWhite  & 0.611 $\pm$ 0.009 (p $<$ 0.010)          & 0.625 $\pm$ 0.009 (p $<$ 0.010)          & 0.634 $\pm$ 0.009 (p $<$ 0.010)          & 0.652 $\pm$ 0.008 (p $<$ 0.010)          & 0.611 $\pm$ 0.008 (p $<$ 0.010)          & 0.627 $\pm$ 0.018 (p $=$ 0.062)          \\
LocalAsian  & 0.637 $\pm$ 0.009 (p $<$ 0.010)          & 0.628 $\pm$ 0.009 (p $<$ 0.010)          & 0.638 $\pm$ 0.010 (p $<$ 0.010)          & 0.642 $\pm$ 0.008 (p $<$ 0.010)          & 0.630 $\pm$ 0.009 (p $<$ 0.010)          & 0.635 $\pm$ 0.006 (p $=$ 0.062)          \\
Centralized & 0.617 $\pm$ 0.010 (p $<$ 0.010)          & 0.643 $\pm$ 0.008 (p $<$ 0.010)          & 0.652 $\pm$ 0.010 (p $<$ 0.010)          & 0.656 $\pm$ 0.009 (p $<$ 0.010)          & 0.635 $\pm$ 0.008 (p $<$ 0.010)          & \textbf{0.640 $\pm$ 0.016 (p $=$ 0.125)} \\
Ensemble    & \textbf{0.641 $\pm$ 0.007 (p $=$ 1.000)} & \textbf{0.656 $\pm$ 0.008 (p $=$ 1.000)} & 0.655 $\pm$ 0.008 (p $<$ 0.010)          & \textbf{0.660 $\pm$ 0.008 (p $=$ 1.000)} & 0.635 $\pm$ 0.008 (p $<$ 0.010)          & \textbf{0.649 $\pm$ 0.011 (p $=$ 1.000)} \\
ModelSoup   & 0.627 $\pm$ 0.009 (p $<$ 0.010)          & 0.647 $\pm$ 0.008 (p $<$ 0.010)          & 0.642 $\pm$ 0.008 (p $<$ 0.010)          & 0.645 $\pm$ 0.008 (p $<$ 0.010)          & 0.622 $\pm$ 0.009 (p $<$ 0.010)          & 0.637 $\pm$ 0.011 (p $=$ 0.062)          \\
FedAvg      & 0.631 $\pm$ 0.008 (p $<$ 0.010)          & 0.643 $\pm$ 0.009 (p $<$ 0.010)          & \textbf{0.665 $\pm$ 0.008 (p $=$ 1.000)} & 0.657 $\pm$ 0.008 (p $=$ 0.014)          & \textbf{0.642 $\pm$ 0.009 (p $=$ 1.000)} & \textbf{0.647 $\pm$ 0.013 (p $=$ 0.625)} \\
\end{tabular}
}

    \label{tab:realistic_patch_accuracy_std_Full}
\end{table}

%  Test set AsianWhite

\begin{table}[H]
    \centering
    \caption{\small Performance of the patch classifiers trained on the population-based heterogeneous setting with 2 clients and evaluated on the population-based test set, in terms of the AUC-ROC One-vs-One (mean $\pm$ covariance). Bold values indicate the best-performing models, whose differences in performance are not statistically significant (Wilcoxon Signed Rank test p-values greater than 0.1).}
    \caption*{\textbf{Population-based setting - Patch Clfs. - Population-based AUC-ROC OvO}}
    \resizebox{\textwidth}{!}{
\begin{tabular}{lcccccc}
\toprule
Model       & Fold 1                                   & Fold 2                                   & Fold 3                                   & Fold 4                                   & Fold 5                                   & CV                                  \\
\midrule
LocalWhite  & 0.879 $\pm$ 0.005 (p $<$ 0.010)          & 0.884 $\pm$ 0.005 (p $<$ 0.010)          & 0.894 $\pm$ 0.005 (p $<$ 0.010)          & 0.897 $\pm$ 0.005 (p $<$ 0.010)          & 0.889 $\pm$ 0.005 (p $<$ 0.010)          & 0.888 $\pm$ 0.008 (p $=$ 0.062)          \\
LocalAsian  & 0.893 $\pm$ 0.005 (p $<$ 0.010)          & 0.900 $\pm$ 0.004 (p $<$ 0.010)          & 0.904 $\pm$ 0.005 (p $<$ 0.010)          & 0.904 $\pm$ 0.005 (p $<$ 0.010)          & 0.896 $\pm$ 0.005 (p $<$ 0.010)          & 0.899 $\pm$ 0.005 (p $=$ 0.062)          \\
Centralized & \textbf{0.898 $\pm$ 0.005 (p $=$ 1.000)} & \textbf{0.908 $\pm$ 0.004 (p $=$ 1.000)} & \textbf{0.911 $\pm$ 0.004 (p $=$ 1.000)} & \textbf{0.907 $\pm$ 0.005 (p $=$ 1.000)} & 0.901 $\pm$ 0.004 (p $=$ 0.011)          & \textbf{0.905 $\pm$ 0.005 (p $=$ 1.000)} \\
Ensemble    & 0.885 $\pm$ 0.005 (p $<$ 0.010)          & 0.897 $\pm$ 0.005 (p $<$ 0.010)          & 0.895 $\pm$ 0.005 (p $<$ 0.010)          & 0.902 $\pm$ 0.004 (p $<$ 0.010)          & 0.891 $\pm$ 0.006 (p $<$ 0.010)          & 0.894 $\pm$ 0.007 (p $=$ 0.062)          \\
ModelSoup   & 0.890 $\pm$ 0.004 (p $<$ 0.010)          & 0.902 $\pm$ 0.005 (p $<$ 0.010)          & 0.906 $\pm$ 0.004 (p $<$ 0.010)          & 0.904 $\pm$ 0.005 (p $<$ 0.010)          & 0.897 $\pm$ 0.005 (p $<$ 0.010)          & 0.899 $\pm$ 0.006 (p $=$ 0.062)          \\
FedAvg      & 0.895 $\pm$ 0.005 (p $<$ 0.010)          & 0.906 $\pm$ 0.005 (p $<$ 0.010)          & \textbf{0.910 $\pm$ 0.005 (p $=$ 0.244)} & 0.905 $\pm$ 0.005 (p $<$ 0.010)          & \textbf{0.902 $\pm$ 0.004 (p $=$ 1.000)} & \textbf{0.904 $\pm$ 0.005 (p $=$ 0.125)} \\
\end{tabular}
}

    \label{tab:realistic_patch_roc_auc_ovo_std_AsianWhite}
\end{table}

\begin{table}[H]
    \centering
    \caption{\small Performance of the patch classifiers trained on the population-based heterogeneous setting with 2 clients and evaluated on the population-based test set, in terms of the AUC-ROC One-vs-Rest (mean $\pm$ covariance). Bold values indicate the best-performing models, whose differences in performance are not statistically significant (Wilcoxon Signed Rank test p-values greater than 0.1).}
    \caption*{\textbf{Population-based setting - Patch Clfs. - Population-based AUC-ROC OvR}}
    \resizebox{\textwidth}{!}{
\begin{tabular}{lcccccc}
\toprule
Model       & Fold 1                                   & Fold 2                                   & Fold 3                                   & Fold 4                                   & Fold 5                                   & CV                                  \\
\midrule
LocalWhite  & 0.809 $\pm$ 0.007 (p $<$ 0.010)          & 0.830 $\pm$ 0.006 (p $<$ 0.010)          & 0.830 $\pm$ 0.007 (p $<$ 0.010)          & 0.838 $\pm$ 0.007 (p $<$ 0.010)          & 0.819 $\pm$ 0.007 (p $<$ 0.010)          & 0.825 $\pm$ 0.012 (p $=$ 0.062)          \\
LocalAsian  & 0.836 $\pm$ 0.006 (p $<$ 0.010)          & 0.846 $\pm$ 0.006 (p $<$ 0.010)          & 0.838 $\pm$ 0.007 (p $<$ 0.010)          & 0.853 $\pm$ 0.006 (p $<$ 0.010)          & 0.840 $\pm$ 0.007 (p $<$ 0.010)          & 0.843 $\pm$ 0.007 (p $=$ 0.062)          \\
Centralized & \textbf{0.848 $\pm$ 0.006 (p $=$ 1.000)} & 0.861 $\pm$ 0.006 (p $<$ 0.010)          & 0.850 $\pm$ 0.006 (p $<$ 0.010)          & 0.847 $\pm$ 0.007 (p $<$ 0.010)          & 0.840 $\pm$ 0.006 (p $<$ 0.010)          & \textbf{0.849 $\pm$ 0.007 (p $=$ 0.125)} \\
Ensemble    & 0.842 $\pm$ 0.006 (p $<$ 0.010)          & 0.857 $\pm$ 0.007 (p $<$ 0.010)          & 0.834 $\pm$ 0.008 (p $<$ 0.010)          & \textbf{0.854 $\pm$ 0.006 (p $=$ 0.145)} & 0.841 $\pm$ 0.008 (p $<$ 0.010)          & \textbf{0.846 $\pm$ 0.010 (p $=$ 0.125)} \\
ModelSoup   & 0.835 $\pm$ 0.006 (p $<$ 0.010)          & 0.855 $\pm$ 0.006 (p $<$ 0.010)          & 0.843 $\pm$ 0.007 (p $<$ 0.010)          & \textbf{0.856 $\pm$ 0.006 (p $=$ 1.000)} & 0.840 $\pm$ 0.006 (p $<$ 0.010)          & \textbf{0.845 $\pm$ 0.009 (p $=$ 0.125)} \\
FedAvg      & 0.844 $\pm$ 0.007 (p $<$ 0.010)          & \textbf{0.864 $\pm$ 0.005 (p $=$ 1.000)} & \textbf{0.857 $\pm$ 0.006 (p $=$ 1.000)} & \textbf{0.854 $\pm$ 0.006 (p $=$ 0.268)} & \textbf{0.854 $\pm$ 0.006 (p $=$ 1.000)} & \textbf{0.855 $\pm$ 0.006 (p $=$ 1.000)} \\
\end{tabular}
}

    \label{tab:realistic_patch_roc_auc_AUC-ROC OvR_std_AsianWhite}
\end{table}

\begin{table}[H]
    \centering
    \caption{\small Performance of the patch classifiers trained on the population-based heterogeneous setting with 2 clients and evaluated population-based the test set, in terms of the accuracy (mean $\pm$ covariance). Bold values indicate the best-performing models, whose differences in performance are not statistically significant (Wilcoxon Signed Rank test p-values greater than 0.1).}
    \caption*{\textbf{Population-based setting (2 clients) - Patch Clfs. - Population-based Accuracy}}
    \resizebox{\textwidth}{!}{
\begin{tabular}{lcccccc}
\toprule
Model       & Fold 1                                   & Fold 2                                   & Fold 3                                   & Fold 4                                   & Fold 5                                   & CV                                  \\
\midrule
LocalWhite  & 0.605 $\pm$ 0.009 (p $<$ 0.010)          & 0.624 $\pm$ 0.010 (p $<$ 0.010)          & 0.631 $\pm$ 0.010 (p $<$ 0.010)          & 0.645 $\pm$ 0.010 (p $<$ 0.010)          & 0.610 $\pm$ 0.009 (p $<$ 0.010)          & 0.623 $\pm$ 0.016 (p $=$ 0.062)          \\
LocalAsian  & 0.629 $\pm$ 0.011 (p $=$ 0.010)          & 0.628 $\pm$ 0.008 (p $<$ 0.010)          & 0.644 $\pm$ 0.010 (p $<$ 0.010)          & 0.638 $\pm$ 0.010 (p $<$ 0.010)          & 0.618 $\pm$ 0.010 (p $<$ 0.010)          & 0.632 $\pm$ 0.010 (p $=$ 0.062)          \\
Centralized & 0.611 $\pm$ 0.010 (p $<$ 0.010)          & 0.635 $\pm$ 0.010 (p $<$ 0.010)          & 0.651 $\pm$ 0.009 (p $<$ 0.010)          & \textbf{0.654 $\pm$ 0.010 (p $=$ 1.000)} & 0.625 $\pm$ 0.010 (p $<$ 0.010)          & \textbf{0.635 $\pm$ 0.017 (p $=$ 0.178)} \\
Ensemble    & \textbf{0.634 $\pm$ 0.011 (p $=$ 1.000)} & \textbf{0.655 $\pm$ 0.010 (p $=$ 1.000)} & 0.659 $\pm$ 0.010 (p $<$ 0.010)          & \textbf{0.652 $\pm$ 0.011 (p $=$ 0.409)} & 0.626 $\pm$ 0.011 (p $<$ 0.010)          & \textbf{0.645 $\pm$ 0.015 (p $=$ 1.000)} \\
ModelSoup   & 0.622 $\pm$ 0.009 (p $<$ 0.010)          & 0.651 $\pm$ 0.011 (p $<$ 0.010)          & 0.651 $\pm$ 0.011 (p $<$ 0.010)          & 0.642 $\pm$ 0.009 (p $<$ 0.010)          & 0.624 $\pm$ 0.009 (p $<$ 0.010)          & 0.638 $\pm$ 0.015 (p $=$ 0.062)          \\
FedAvg      & 0.621 $\pm$ 0.010 (p $<$ 0.010)          & 0.640 $\pm$ 0.010 (p $<$ 0.010)          & \textbf{0.668 $\pm$ 0.011 (p $=$ 1.000)} & 0.648 $\pm$ 0.010 (p $<$ 0.010)          & \textbf{0.633 $\pm$ 0.010 (p $=$ 1.000)} & \textbf{0.642 $\pm$ 0.016 (p $=$ 0.625)} \\
\end{tabular}
}

    \label{tab:realistic_patch_accuracy_std_AsianWhite}
\end{table}

% Asian population

\begin{table}[H]
    \centering
    \caption{\small Performance of the patch classifiers trained on the population-based heterogeneous setting with 2 clients and evaluated the Asian population-based test set, in terms of the AUC-ROC One-vs-One (mean $\pm$ covariance). Bold values indicate the best-performing models, whose differences in performance are not statistically significant (Wilcoxon Signed Rank test p-values greater than 0.1).}
    \caption*{\textbf{Population-based setting - Patch Clfs. - Asian population-based AUC-ROC OvO}}
    \resizebox{\textwidth}{!}{
\begin{tabular}{lcccccc}
\toprule
Model       & Fold 1                                   & Fold 2                                   & Fold 3                                   & Fold 4                                   & Fold 5                                   & CV                                  \\
\midrule
LocalWhite  & 0.891 $\pm$ 0.006 (p $<$ 0.010)          & 0.893 $\pm$ 0.007 (p $<$ 0.010)          & 0.903 $\pm$ 0.006 (p $<$ 0.010)          & 0.912 $\pm$ 0.006 (p $<$ 0.010)          & 0.899 $\pm$ 0.006 (p $<$ 0.010)          & 0.899 $\pm$ 0.009 (p $=$ 0.062)          \\
LocalAsian  & 0.909 $\pm$ 0.006 (p $<$ 0.010)          & 0.920 $\pm$ 0.005 (p $<$ 0.010)          & 0.906 $\pm$ 0.006 (p $<$ 0.010)          & 0.925 $\pm$ 0.005 (p $<$ 0.010)          & 0.906 $\pm$ 0.006 (p $<$ 0.010)          & 0.913 $\pm$ 0.009 (p $=$ 0.062)          \\
Centralized & \textbf{0.920 $\pm$ 0.005 (p $=$ 1.000)} & \textbf{0.928 $\pm$ 0.005 (p $=$ 1.000)} & \textbf{0.922 $\pm$ 0.005 (p $=$ 0.997)} & \textbf{0.928 $\pm$ 0.004 (p $=$ 1.000)} & 0.915 $\pm$ 0.005 (p $<$ 0.010)          & \textbf{0.923 $\pm$ 0.006 (p $=$ 1.000)} \\
Ensemble    & 0.893 $\pm$ 0.006 (p $<$ 0.010)          & 0.911 $\pm$ 0.005 (p $<$ 0.010)          & 0.895 $\pm$ 0.006 (p $<$ 0.010)          & 0.918 $\pm$ 0.006 (p $<$ 0.010)          & 0.897 $\pm$ 0.006 (p $<$ 0.010)          & 0.903 $\pm$ 0.011 (p $=$ 0.062)          \\
ModelSoup   & 0.907 $\pm$ 0.006 (p $<$ 0.010)          & 0.916 $\pm$ 0.005 (p $<$ 0.010)          & 0.908 $\pm$ 0.006 (p $<$ 0.010)          & 0.922 $\pm$ 0.005 (p $<$ 0.010)          & 0.905 $\pm$ 0.005 (p $<$ 0.010)          & 0.912 $\pm$ 0.007 (p $=$ 0.062)          \\
FedAvg      & 0.914 $\pm$ 0.005 (p $<$ 0.010)          & 0.925 $\pm$ 0.005 (p $<$ 0.010)          & \textbf{0.923 $\pm$ 0.005 (p $=$ 1.000)} & \textbf{0.928 $\pm$ 0.004 (p $=$ 1.000)} & \textbf{0.919 $\pm$ 0.005 (p $=$ 1.000)} & \textbf{0.922 $\pm$ 0.006 (p $=$ 0.812)} \\
\end{tabular}
}

    \label{tab:realistic_patch_roc_auc_ovo_std_Asian-SWOR}
\end{table}

\begin{table}[H]
    \centering
    \caption{\small Performance of the patch classifiers trained on the population-based heterogeneous setting with 2 clients and evaluated the Asian population-based test set, in terms of the AUC-ROC One-vs-Rest (mean $\pm$ covariance). Bold values indicate the best-performing models, whose differences in performance are not statistically significant (Wilcoxon Signed Rank test p-values greater than 0.1).}
    \caption*{\textbf{Population-based setting - Patch Clfs. - Asian population-based AUC-ROC OvR}}
    \resizebox{\textwidth}{!}{
\begin{tabular}{lcccccc}
\toprule
Model       & Fold 1                                   & Fold 2                                   & Fold 3                                   & Fold 4                                   & Fold 5                                   & CV                                  \\
\midrule
LocalWhite  & 0.815 $\pm$ 0.009 (p $<$ 0.010)          & 0.829 $\pm$ 0.010 (p $<$ 0.010)          & 0.827 $\pm$ 0.009 (p $<$ 0.010)          & 0.852 $\pm$ 0.009 (p $<$ 0.010)          & 0.818 $\pm$ 0.009 (p $<$ 0.010)          & 0.828 $\pm$ 0.015 (p $=$ 0.062)          \\
LocalAsian  & 0.840 $\pm$ 0.010 (p $<$ 0.010)          & 0.859 $\pm$ 0.009 (p $<$ 0.010)          & 0.817 $\pm$ 0.011 (p $<$ 0.010)          & 0.865 $\pm$ 0.009 (p $<$ 0.010)          & 0.840 $\pm$ 0.009 (p $<$ 0.010)          & 0.844 $\pm$ 0.019 (p $=$ 0.062)          \\
Centralized & \textbf{0.866 $\pm$ 0.008 (p $=$ 1.000)} & \textbf{0.883 $\pm$ 0.007 (p $=$ 1.000)} & 0.845 $\pm$ 0.010 (p $<$ 0.010)          & 0.864 $\pm$ 0.008 (p $<$ 0.010)          & 0.839 $\pm$ 0.009 (p $<$ 0.010)          & \textbf{0.859 $\pm$ 0.018 (p $=$ 0.312)} \\
Ensemble    & 0.836 $\pm$ 0.009 (p $<$ 0.010)          & 0.864 $\pm$ 0.008 (p $<$ 0.010)          & 0.808 $\pm$ 0.010 (p $<$ 0.010)          & 0.864 $\pm$ 0.010 (p $<$ 0.010)          & 0.836 $\pm$ 0.010 (p $<$ 0.010)          & 0.842 $\pm$ 0.023 (p $=$ 0.062)          \\
ModelSoup   & 0.844 $\pm$ 0.009 (p $<$ 0.010)          & 0.859 $\pm$ 0.008 (p $<$ 0.010)          & 0.830 $\pm$ 0.010 (p $<$ 0.010)          & 0.868 $\pm$ 0.009 (p $<$ 0.010)          & 0.833 $\pm$ 0.010 (p $<$ 0.010)          & 0.847 $\pm$ 0.016 (p $=$ 0.062)          \\
FedAvg      & 0.857 $\pm$ 0.008 (p $<$ 0.010)          & 0.880 $\pm$ 0.007 (p $<$ 0.010)          & \textbf{0.854 $\pm$ 0.010 (p $=$ 1.000)} & \textbf{0.873 $\pm$ 0.008 (p $=$ 1.000)} & \textbf{0.862 $\pm$ 0.009 (p $=$ 1.000)} & \textbf{0.866 $\pm$ 0.011 (p $=$ 1.000)} \\
\end{tabular}
}

    \label{tab:realistic_patch_roc_auc_AUC-ROC OvR_std_Asian-SWOR}
\end{table}

\begin{table}[H]
    \centering
    \caption{\small Performance of the patch classifiers trained on the population-based heterogeneous setting with 2 clients and evaluated the Asian population-based test set, in terms of the accuracy (mean $\pm$ covariance). Bold values indicate the best-performing models, whose differences in performance are not statistically significant (Wilcoxon Signed Rank test p-values greater than 0.1).}
    \caption*{\textbf{Population-based setting (2 clients) - Patch Clfs. - Asian population-based Accuracy}}
    \resizebox{\textwidth}{!}{
\begin{tabular}{lcccccc}
\toprule
Model       & Fold 1                                   & Fold 2                                   & Fold 3                                   & Fold 4                                   & Fold 5                                   & CV                                  \\
\midrule
LocalWhite  & 0.595 $\pm$ 0.015 (p $<$ 0.010)          & 0.617 $\pm$ 0.013 (p $<$ 0.010)          & 0.635 $\pm$ 0.013 (p $<$ 0.010)          & 0.662 $\pm$ 0.014 (p $<$ 0.010)          & 0.600 $\pm$ 0.016 (p $<$ 0.010)          & 0.622 $\pm$ 0.027 (p $=$ 0.062)          \\
LocalAsian  & 0.641 $\pm$ 0.013 (p $<$ 0.010)          & 0.658 $\pm$ 0.012 (p $<$ 0.010)          & 0.654 $\pm$ 0.012 (p $<$ 0.010)          & 0.662 $\pm$ 0.014 (p $<$ 0.010)          & 0.631 $\pm$ 0.015 (p $<$ 0.010)          & \textbf{0.649 $\pm$ 0.013 (p $=$ 0.178)} \\
Centralized & 0.634 $\pm$ 0.016 (p $<$ 0.010)          & 0.640 $\pm$ 0.015 (p $<$ 0.010)          & 0.660 $\pm$ 0.015 (p $<$ 0.010)          & \textbf{0.680 $\pm$ 0.013 (p $=$ 1.000)} & 0.632 $\pm$ 0.015 (p $<$ 0.010)          & 0.649 $\pm$ 0.020 (p $=$ 0.059)          \\
Ensemble    & \textbf{0.655 $\pm$ 0.014 (p $=$ 1.000)} & \textbf{0.665 $\pm$ 0.014 (p $=$ 1.000)} & 0.666 $\pm$ 0.012 (p $<$ 0.010)          & \textbf{0.680 $\pm$ 0.015 (p $=$ 1.000)} & 0.624 $\pm$ 0.014 (p $<$ 0.010)          & \textbf{0.658 $\pm$ 0.021 (p $=$ 0.812)} \\
ModelSoup   & 0.635 $\pm$ 0.015 (p $<$ 0.010)          & \textbf{0.664 $\pm$ 0.013 (p $=$ 0.807)} & 0.667 $\pm$ 0.014 (p $<$ 0.010)          & 0.672 $\pm$ 0.013 (p $<$ 0.010)          & 0.631 $\pm$ 0.014 (p $<$ 0.010)          & \textbf{0.654 $\pm$ 0.018 (p $=$ 0.188)} \\
FedAvg      & 0.641 $\pm$ 0.015 (p $<$ 0.010)          & 0.656 $\pm$ 0.016 (p $<$ 0.010)          & \textbf{0.689 $\pm$ 0.015 (p $=$ 1.000)} & \textbf{0.678 $\pm$ 0.012 (p $=$ 0.517)} & \textbf{0.646 $\pm$ 0.014 (p $=$ 1.000)} & \textbf{0.662 $\pm$ 0.022 (p $=$ 1.000)} \\
\end{tabular}
}

    \label{tab:realistic_patch_accuracy_std_Asian-SWOR}
\end{table}

% White population

\begin{table}[H]
    \centering
    \caption{\small Performance of the patch classifiers trained on the population-based heterogeneous setting with 2 clients and evaluated on the White population-based test set, in terms of the AUC-ROC One-vs-One (mean $\pm$ covariance). Bold values indicate the best-performing models, whose differences in performance are not statistically significant (Wilcoxon Signed Rank test p-values greater than 0.1).}
    \caption*{\textbf{Population-based setting - Patch Clfs. - White population-based AUC-ROC OvO}}
    \resizebox{\textwidth}{!}{
\begin{tabular}{lcccccc}
\toprule
Model       & Fold 1                                   & Fold 2                                   & Fold 3                                   & Fold 4                                   & Fold 5                                   & CV                                  \\
\midrule
LocalWhite  & 0.868 $\pm$ 0.008 (p $<$ 0.010)          & 0.876 $\pm$ 0.008 (p $<$ 0.010)          & 0.888 $\pm$ 0.007 (p $<$ 0.010)          & 0.883 $\pm$ 0.009 (p $<$ 0.010)          & 0.879 $\pm$ 0.007 (p $<$ 0.010)          & 0.879 $\pm$ 0.008 (p $=$ 0.062)          \\
LocalAsian  & 0.877 $\pm$ 0.009 (p $<$ 0.010)          & 0.882 $\pm$ 0.008 (p $<$ 0.010)          & \textbf{0.905 $\pm$ 0.008 (p $=$ 1.000)} & 0.882 $\pm$ 0.008 (p $<$ 0.010)          & \textbf{0.889 $\pm$ 0.009 (p $=$ 0.700)} & \textbf{0.888 $\pm$ 0.011 (p $=$ 0.312)} \\
Centralized & 0.879 $\pm$ 0.008 (p $<$ 0.010)          & 0.888 $\pm$ 0.007 (p $<$ 0.010)          & 0.902 $\pm$ 0.006 (p $<$ 0.010)          & \textbf{0.888 $\pm$ 0.007 (p $=$ 1.000)} & \textbf{0.890 $\pm$ 0.008 (p $=$ 1.000)} & \textbf{0.890 $\pm$ 0.009 (p $=$ 1.000)} \\
Ensemble    & 0.876 $\pm$ 0.007 (p $<$ 0.010)          & 0.883 $\pm$ 0.008 (p $<$ 0.010)          & 0.896 $\pm$ 0.006 (p $<$ 0.010)          & \textbf{0.888 $\pm$ 0.007 (p $=$ 1.000)} & 0.884 $\pm$ 0.007 (p $<$ 0.010)          & 0.886 $\pm$ 0.006 (p $=$ 0.062)          \\
ModelSoup   & 0.874 $\pm$ 0.008 (p $<$ 0.010)          & \textbf{0.892 $\pm$ 0.008 (p $=$ 1.000)} & \textbf{0.904 $\pm$ 0.007 (p $=$ 0.154)} & 0.886 $\pm$ 0.008 (p $<$ 0.010)          & 0.888 $\pm$ 0.008 (p $=$ 0.097)          & \textbf{0.889 $\pm$ 0.010 (p $=$ 0.625)} \\
FedAvg      & \textbf{0.882 $\pm$ 0.007 (p $=$ 1.000)} & 0.887 $\pm$ 0.008 (p $<$ 0.010)          & 0.899 $\pm$ 0.008 (p $<$ 0.010)          & 0.883 $\pm$ 0.008 (p $<$ 0.010)          & 0.888 $\pm$ 0.007 (p $=$ 0.048)          & \textbf{0.888 $\pm$ 0.007 (p $=$ 0.438)} \\
\end{tabular}
}
    \label{tab:realistic_patch_roc_auc_ovo_std_White-SWOR}
\end{table}

\begin{table}[H]
    \centering
    \caption{\small Performance of the patch classifiers trained on the population-based heterogeneous setting with 2 clients and evaluated on the White population-based test set, in terms of the AUC-ROC One-vs-Rest (mean $\pm$ covariance). Bold values indicate the best-performing models, whose differences in performance are not statistically significant (Wilcoxon Signed Rank test p-values greater than 0.1).}
    \caption*{\textbf{Population-based setting - Patch Clfs. - White population-based AUC-ROC OvR}}
    \resizebox{\textwidth}{!}{
\begin{tabular}{lcccccc}
\toprule
Model       & Fold 1                                   & Fold 2                                   & Fold 3                                   & Fold 4                                   & Fold 5                                   & CV                                  \\
\midrule
LocalWhite  & 0.802 $\pm$ 0.011 (p $<$ 0.010)          & 0.829 $\pm$ 0.009 (p $<$ 0.010)          & 0.835 $\pm$ 0.008 (p $<$ 0.010)          & 0.826 $\pm$ 0.011 (p $<$ 0.010)          & 0.820 $\pm$ 0.010 (p $<$ 0.010)          & 0.822 $\pm$ 0.012 (p $=$ 0.062)          \\
LocalAsian  & 0.829 $\pm$ 0.009 (p $<$ 0.010)          & 0.837 $\pm$ 0.008 (p $<$ 0.010)          & \textbf{0.856 $\pm$ 0.010 (p $=$ 0.498)} & 0.840 $\pm$ 0.008 (p $<$ 0.010)          & 0.841 $\pm$ 0.010 (p $<$ 0.010)          & \textbf{0.842 $\pm$ 0.010 (p $=$ 0.125)} \\
Centralized & 0.834 $\pm$ 0.009 (p $<$ 0.010)          & 0.840 $\pm$ 0.008 (p $<$ 0.010)          & 0.855 $\pm$ 0.007 (p $=$ 0.034)          & 0.831 $\pm$ 0.009 (p $<$ 0.010)          & 0.841 $\pm$ 0.009 (p $<$ 0.010)          & \textbf{0.841 $\pm$ 0.009 (p $=$ 0.125)} \\
Ensemble    & \textbf{0.848 $\pm$ 0.008 (p $=$ 1.000)} & 0.852 $\pm$ 0.008 (p $=$ 0.098)          & 0.856 $\pm$ 0.007 (p $=$ 0.039)          & \textbf{0.848 $\pm$ 0.009 (p $=$ 1.000)} & \textbf{0.846 $\pm$ 0.008 (p $=$ 1.000)} & \textbf{0.850 $\pm$ 0.004 (p $=$ 1.000)} \\
ModelSoup   & 0.826 $\pm$ 0.009 (p $<$ 0.010)          & \textbf{0.854 $\pm$ 0.008 (p $=$ 1.000)} & 0.854 $\pm$ 0.009 (p $<$ 0.010)          & 0.844 $\pm$ 0.009 (p $<$ 0.010)          & 0.842 $\pm$ 0.009 (p $<$ 0.010)          & \textbf{0.844 $\pm$ 0.011 (p $=$ 0.125)} \\
FedAvg      & 0.835 $\pm$ 0.008 (p $<$ 0.010)          & 0.848 $\pm$ 0.009 (p $<$ 0.010)          & \textbf{0.858 $\pm$ 0.009 (p $=$ 1.000)} & 0.837 $\pm$ 0.009 (p $<$ 0.010)          & \textbf{0.846 $\pm$ 0.008 (p $=$ 1.000)} & \textbf{0.845 $\pm$ 0.009 (p $=$ 0.312)} \\
\end{tabular}
}

    \label{tab:realistic_patch_roc_auc_AUC-ROC OvR_std_White-SWOR}
\end{table}

\begin{table}[H]
    \centering
    \caption{\small Performance of the patch classifiers trained on the population-based heterogeneous setting with 2 clients and evaluated on the White population-based test set, in terms of the accuracy (mean $\pm$ covariance). Bold values indicate the best-performing models, whose differences in performance are not statistically significant (Wilcoxon Signed Rank test p-values greater than 0.1).}
    \caption*{\textbf{Population-based setting (2 clients) - Patch Clfs. - White population-based Accuracy}}
    \resizebox{\textwidth}{!}{
\begin{tabular}{lcccccc}
\toprule
Model       & Fold 1                                   & Fold 2                                   & Fold 3                                   & Fold 4                                   & Fold 5                                   & CV                                  \\
\midrule
LocalWhite  & \textbf{0.617 $\pm$ 0.013 (p $=$ 1.000)} & 0.628 $\pm$ 0.014 (p $<$ 0.010)          & 0.630 $\pm$ 0.012 (p $<$ 0.010)          & \textbf{0.631 $\pm$ 0.015 (p $=$ 1.000)} & 0.617 $\pm$ 0.015 (p $<$ 0.010)          & \textbf{0.624 $\pm$ 0.006 (p $=$ 0.312)} \\
LocalAsian  & 0.613 $\pm$ 0.016 (p $=$ 0.064)          & 0.599 $\pm$ 0.014 (p $<$ 0.010)          & 0.636 $\pm$ 0.015 (p $<$ 0.010)          & 0.611 $\pm$ 0.013 (p $<$ 0.010)          & 0.610 $\pm$ 0.015 (p $<$ 0.010)          & \textbf{0.615 $\pm$ 0.013 (p $=$ 0.125)} \\
Centralized & 0.590 $\pm$ 0.014 (p $<$ 0.010)          & 0.630 $\pm$ 0.013 (p $<$ 0.010)          & 0.641 $\pm$ 0.014 (p $<$ 0.010)          & \textbf{0.629 $\pm$ 0.014 (p $=$ 0.116)} & 0.617 $\pm$ 0.015 (p $<$ 0.010)          & \textbf{0.621 $\pm$ 0.019 (p $=$ 0.125)} \\
Ensemble    & 0.610 $\pm$ 0.013 (p $<$ 0.010)          & \textbf{0.646 $\pm$ 0.013 (p $=$ 0.608)} & \textbf{0.649 $\pm$ 0.014 (p $=$ 1.000)} & 0.627 $\pm$ 0.013 (p $=$ 0.038)          & \textbf{0.625 $\pm$ 0.014 (p $=$ 1.000)} & \textbf{0.632 $\pm$ 0.014 (p $=$ 1.000)} \\
ModelSoup   & 0.603 $\pm$ 0.015 (p $<$ 0.010)          & \textbf{0.647 $\pm$ 0.013 (p $=$ 1.000)} & 0.636 $\pm$ 0.013 (p $<$ 0.010)          & 0.613 $\pm$ 0.015 (p $<$ 0.010)          & 0.611 $\pm$ 0.015 (p $<$ 0.010)          & 0.622 $\pm$ 0.016 (p $=$ 0.062)          \\
FedAvg      & 0.608 $\pm$ 0.014 (p $<$ 0.010)          & 0.623 $\pm$ 0.015 (p $<$ 0.010)          & 0.646 $\pm$ 0.015 (p $=$ 0.078)          & 0.616 $\pm$ 0.013 (p $<$ 0.010)          & 0.618 $\pm$ 0.014 (p $<$ 0.010)          & 0.623 $\pm$ 0.014 (p $=$ 0.062)          \\
\end{tabular}
}

    \label{tab:realistic_patch_accuracy_std_White-SWOR}
\end{table}

\paragraph{Whole image classifier}
\label{app:sec:pop_based_set_whole_img}

% Overall test set

\begin{table}[H]
    \centering
    \caption{\small Performance of the whole image classifiers trained on the population-based heterogeneous setting with 2 clients and evaluated on the original test set, in terms of the AUC-ROC (mean $\pm$ covariance). Bold values indicate the best-performing models, whose differences in performance are not statistically significant (Wilcoxon Signed Rank test p-values greater than 0.1).}
    \caption*{\textbf{Population-based setting - Whole Image Clfs. - Overall AUC-ROC.}}
    \resizebox{\textwidth}{!}{
\begin{tabular}{lcccccc}
\toprule
Model       & Fold 1                                   & Fold 2                                   & Fold 3                                   & Fold 4                                   & Fold 5                                   & CV                                  \\
\midrule
LocalWhite  & 0.748 $\pm$ 0.031 (p $<$ 0.010)          & 0.735 $\pm$ 0.031 (p $<$ 0.010)          & 0.663 $\pm$ 0.034 (p $<$ 0.010)          & 0.720 $\pm$ 0.032 (p $<$ 0.010)          & 0.710 $\pm$ 0.033 (p $<$ 0.010)          & 0.712 $\pm$ 0.034 (p $=$ 0.062)          \\
LocalAsian  & 0.691 $\pm$ 0.033 (p $<$ 0.010)          & 0.738 $\pm$ 0.033 (p $<$ 0.010)          & \textbf{0.753 $\pm$ 0.032 (p $=$ 1.000)} & 0.700 $\pm$ 0.035 (p $<$ 0.010)          & 0.717 $\pm$ 0.028 (p $<$ 0.010)          & \textbf{0.720 $\pm$ 0.027 (p $=$ 0.125)} \\
Centralized & 0.718 $\pm$ 0.032 (p $<$ 0.010)          & 0.705 $\pm$ 0.037 (p $<$ 0.010)          & 0.737 $\pm$ 0.033 (p $<$ 0.010)          & 0.737 $\pm$ 0.030 (p $<$ 0.010)          & 0.730 $\pm$ 0.030 (p $<$ 0.010)          & 0.726 $\pm$ 0.014 (p $=$ 0.062)          \\
Ensemble    & 0.752 $\pm$ 0.030 (p $=$ 0.013)          & 0.757 $\pm$ 0.033 (p $=$ 0.068)          & 0.736 $\pm$ 0.032 (p $<$ 0.010)          & 0.729 $\pm$ 0.036 (p $<$ 0.010)          & 0.734 $\pm$ 0.032 (p $<$ 0.010)          & 0.741 $\pm$ 0.013 (p $=$ 0.062)          \\
ModelSoup   & 0.719 $\pm$ 0.039 (p $<$ 0.010)          & 0.737 $\pm$ 0.036 (p $<$ 0.010)          & 0.734 $\pm$ 0.033 (p $<$ 0.010)          & 0.717 $\pm$ 0.036 (p $<$ 0.010)          & 0.730 $\pm$ 0.028 (p $<$ 0.010)          & 0.726 $\pm$ 0.010 (p $=$ 0.062)          \\
FedAvg      & \textbf{0.764 $\pm$ 0.029 (p $=$ 1.000)} & \textbf{0.764 $\pm$ 0.031 (p $=$ 1.000)} & 0.746 $\pm$ 0.029 (p $=$ 0.069)          & \textbf{0.789 $\pm$ 0.031 (p $=$ 1.000)} & \textbf{0.755 $\pm$ 0.026 (p $=$ 1.000)} & \textbf{0.764 $\pm$ 0.017 (p $=$ 1.000)} \\
\end{tabular}
}

    \label{tab:realitic_roc_auc_std_Full}
\end{table}

\subsubsection{Strongly heterogeneous setting - 2 clients}

\paragraph{Patch classifier}
\label{app:sec:strong_het_2_clients_patch}

% Overall test set

\begin{table}[H]
    \centering
    \caption{\small Performance of the patch classifiers trained on the strongly heterogeneous setting with 2 clients and evaluated on the original test set, in terms of the AUC-ROC One-vs-One (mean $\pm$ covariance). Bold values indicate the best-performing models, whose differences in performance are not statistically significant (Wilcoxon Signed Rank test p-values greater than 0.1).}
    \caption*{\textbf{Strong setting (2 clients) - Patch Clfs. - Overall AUC-ROC OvO}}
    \resizebox{\textwidth}{!}{
\begin{tabular}{lcccccc}
\toprule
Model       & Fold 1                                   & Fold 2                                   & Fold 3                                   & Fold 4                                   & Fold 5                                   & CV                                  \\
\midrule
LocalLow    & 0.898 $\pm$ 0.004 (p $<$ 0.010)          & 0.905 $\pm$ 0.003 (p $<$ 0.010)          & 0.907 $\pm$ 0.004 (p $<$ 0.010)          & 0.890 $\pm$ 0.004 (p $<$ 0.010)          & 0.903 $\pm$ 0.004 (p $<$ 0.010)          & 0.901 $\pm$ 0.007 (p $=$ 0.062)          \\
LocalHigh   & 0.900 $\pm$ 0.005 (p $<$ 0.010)          & 0.903 $\pm$ 0.004 (p $<$ 0.010)          & 0.904 $\pm$ 0.004 (p $<$ 0.010)          & 0.901 $\pm$ 0.004 (p $<$ 0.010)          & 0.902 $\pm$ 0.004 (p $<$ 0.010)          & 0.902 $\pm$ 0.001 (p $=$ 0.062)          \\
Centralized & \textbf{0.919 $\pm$ 0.004 (p $=$ 1.000)} & \textbf{0.917 $\pm$ 0.004 (p $=$ 1.000)} & \textbf{0.922 $\pm$ 0.005 (p $=$ 1.000)} & \textbf{0.920 $\pm$ 0.004 (p $=$ 1.000)} & \textbf{0.912 $\pm$ 0.004 (p $=$ 0.431)} & \textbf{0.918 $\pm$ 0.004 (p $=$ 1.000)} \\
Ensemble    & 0.895 $\pm$ 0.004 (p $<$ 0.010)          & 0.898 $\pm$ 0.005 (p $<$ 0.010)          & 0.902 $\pm$ 0.004 (p $<$ 0.010)          & 0.893 $\pm$ 0.004 (p $<$ 0.010)          & 0.900 $\pm$ 0.004 (p $<$ 0.010)          & 0.898 $\pm$ 0.003 (p $=$ 0.062)          \\
ModelSoup   & 0.907 $\pm$ 0.004 (p $<$ 0.010)          & 0.906 $\pm$ 0.004 (p $<$ 0.010)          & 0.910 $\pm$ 0.004 (p $<$ 0.010)          & 0.899 $\pm$ 0.004 (p $<$ 0.010)          & 0.911 $\pm$ 0.004 (p $=$ 0.025)          & \textbf{0.907 $\pm$ 0.005 (p $=$ 0.125)} \\
FedAvg      & 0.915 $\pm$ 0.004 (p $<$ 0.010)          & 0.916 $\pm$ 0.004 (p $<$ 0.010)          & 0.921 $\pm$ 0.004 (p $=$ 0.013)          & 0.913 $\pm$ 0.004 (p $<$ 0.010)          & 0.910 $\pm$ 0.004 (p $<$ 0.010)          & 0.915 $\pm$ 0.004 (p $=$ 0.062)          \\
FedProx     & 0.915 $\pm$ 0.003 (p $<$ 0.010)          & 0.913 $\pm$ 0.004 (p $<$ 0.010)          & 0.913 $\pm$ 0.005 (p $<$ 0.010)          & 0.909 $\pm$ 0.005 (p $<$ 0.010)          & 0.903 $\pm$ 0.005 (p $<$ 0.010)          & 0.911 $\pm$ 0.005 (p $=$ 0.062)          \\
SCAFFOLD    & 0.917 $\pm$ 0.004 (p $<$ 0.010)          & 0.914 $\pm$ 0.004 (p $<$ 0.010)          & 0.913 $\pm$ 0.004 (p $<$ 0.010)          & 0.914 $\pm$ 0.004 (p $<$ 0.010)          & \textbf{0.913 $\pm$ 0.004 (p $=$ 1.000)} & \textbf{0.914 $\pm$ 0.002 (p $=$ 0.125)} \\
\end{tabular}
}

    \label{tab:strong_patch_roc_auc_ovo_std_Full}
\end{table}

\begin{table}[H]
    \centering
    \caption{\small Performance of the patch classifiers trained on the strongly heterogeneous setting with 2 clients and evaluated on the original test set, in terms of the AUC-ROC One-vs-Rest (mean $\pm$ covariance). Bold values indicate the best-performing models, whose differences in performance are not statistically significant (Wilcoxon Signed Rank test p-values greater than 0.1).}
    \caption*{\textbf{Strong setting (2 clients) - Patch Clfs. - Overall AUC-ROC OvR}}
    \resizebox{\textwidth}{!}{
\begin{tabular}{lcccccc}
\toprule
Model       & Fold 1                                   & Fold 2                                   & Fold 3                                   & Fold 4                                   & Fold 5                                   & CV                                  \\
\midrule
LocalLow    & 0.850 $\pm$ 0.005 (p $<$ 0.010)          & 0.864 $\pm$ 0.004 (p $<$ 0.010)          & 0.854 $\pm$ 0.006 (p $<$ 0.010)          & 0.835 $\pm$ 0.006 (p $<$ 0.010)          & 0.843 $\pm$ 0.005 (p $<$ 0.010)          & 0.849 $\pm$ 0.011 (p $=$ 0.062)          \\
LocalHigh   & 0.851 $\pm$ 0.006 (p $<$ 0.010)          & 0.860 $\pm$ 0.006 (p $<$ 0.010)          & 0.848 $\pm$ 0.005 (p $<$ 0.010)          & 0.855 $\pm$ 0.005 (p $<$ 0.010)          & 0.859 $\pm$ 0.005 (p $<$ 0.010)          & 0.855 $\pm$ 0.005 (p $=$ 0.062)          \\
Centralized & 0.875 $\pm$ 0.006 (p $<$ 0.010)          & 0.878 $\pm$ 0.005 (p $<$ 0.010)          & \textbf{0.877 $\pm$ 0.006 (p $=$ 1.000)} & \textbf{0.874 $\pm$ 0.005 (p $=$ 1.000)} & 0.870 $\pm$ 0.005 (p $<$ 0.010)          & \textbf{0.874 $\pm$ 0.003 (p $=$ 0.312)} \\
Ensemble    & 0.854 $\pm$ 0.006 (p $<$ 0.010)          & 0.862 $\pm$ 0.006 (p $<$ 0.010)          & 0.856 $\pm$ 0.006 (p $<$ 0.010)          & 0.853 $\pm$ 0.005 (p $<$ 0.010)          & 0.853 $\pm$ 0.006 (p $<$ 0.010)          & 0.856 $\pm$ 0.004 (p $=$ 0.062)          \\
ModelSoup   & 0.858 $\pm$ 0.006 (p $<$ 0.010)          & 0.865 $\pm$ 0.005 (p $<$ 0.010)          & 0.862 $\pm$ 0.005 (p $<$ 0.010)          & 0.846 $\pm$ 0.005 (p $<$ 0.010)          & 0.856 $\pm$ 0.005 (p $<$ 0.010)          & 0.857 $\pm$ 0.007 (p $=$ 0.062)          \\
FedAvg      & 0.872 $\pm$ 0.005 (p $<$ 0.010)          & \textbf{0.881 $\pm$ 0.005 (p $=$ 1.000)} & \textbf{0.877 $\pm$ 0.006 (p $=$ 1.000)} & 0.872 $\pm$ 0.005 (p $=$ 0.015)          & 0.871 $\pm$ 0.005 (p $<$ 0.010)          & \textbf{0.874 $\pm$ 0.004 (p $=$ 0.812)} \\
FedProx     & \textbf{0.878 $\pm$ 0.004 (p $=$ 0.171)} & 0.877 $\pm$ 0.005 (p $<$ 0.010)          & \textbf{0.877 $\pm$ 0.006 (p $=$ 1.000)} & \textbf{0.874 $\pm$ 0.006 (p $=$ 1.000)} & 0.869 $\pm$ 0.006 (p $<$ 0.010)          & \textbf{0.875 $\pm$ 0.004 (p $=$ 1.000)} \\
SCAFFOLD    & \textbf{0.879 $\pm$ 0.005 (p $=$ 1.000)} & 0.877 $\pm$ 0.005 (p $<$ 0.010)          & 0.873 $\pm$ 0.006 (p $<$ 0.010)          & \textbf{0.874 $\pm$ 0.005 (p $=$ 1.000)} & \textbf{0.874 $\pm$ 0.005 (p $=$ 1.000)} & \textbf{0.875 $\pm$ 0.003 (p $=$ 1.000)} \\
\end{tabular}
}
    \label{tab:strong_patch_roc_auc_AUC-ROC OvR_std_Full}
\end{table}

\begin{table}[H]
    \centering
    \caption{\small Performance of the patch classifiers trained on the strongly heterogeneous setting with 2 clients and evaluated on the original test set, in terms of the accuracy (mean $\pm$ covariance). Bold values indicate the best-performing models, whose differences in performance are not statistically significant (Wilcoxon Signed Rank test p-values greater than 0.1).}
    \caption*{\textbf{Strong setting (2 clients) - Patch Clfs. - Overall Accuracy}}
    \resizebox{\textwidth}{!}{
\begin{tabular}{lcccccc}
\toprule
Model       & Fold 1                                   & Fold 2                                   & Fold 3                                   & Fold 4                                   & Fold 5                                   & CV                                  \\
\midrule
LocalLow    & 0.637 $\pm$ 0.008 (p $<$ 0.010)          & 0.637 $\pm$ 0.009 (p $<$ 0.010)          & 0.661 $\pm$ 0.010 (p $<$ 0.010)          & 0.610 $\pm$ 0.009 (p $<$ 0.010)          & 0.647 $\pm$ 0.008 (p $<$ 0.010)          & 0.638 $\pm$ 0.019 (p $=$ 0.062)          \\
LocalHigh   & 0.655 $\pm$ 0.010 (p $<$ 0.010)          & 0.604 $\pm$ 0.009 (p $<$ 0.010)          & 0.622 $\pm$ 0.009 (p $<$ 0.010)          & 0.635 $\pm$ 0.008 (p $<$ 0.010)          & 0.623 $\pm$ 0.009 (p $<$ 0.010)          & 0.627 $\pm$ 0.019 (p $=$ 0.062)          \\
Centralized & \textbf{0.678 $\pm$ 0.008 (p $=$ 1.000)} & 0.674 $\pm$ 0.008 (p $<$ 0.010)          & \textbf{0.687 $\pm$ 0.009 (p $=$ 1.000)} & \textbf{0.674 $\pm$ 0.008 (p $=$ 0.213)} & \textbf{0.667 $\pm$ 0.009 (p $=$ 1.000)} & \textbf{0.676 $\pm$ 0.007 (p $=$ 1.000)} \\
Ensemble    & 0.665 $\pm$ 0.008 (p $<$ 0.010)          & 0.633 $\pm$ 0.009 (p $<$ 0.010)          & 0.661 $\pm$ 0.008 (p $<$ 0.010)          & 0.644 $\pm$ 0.008 (p $<$ 0.010)          & 0.656 $\pm$ 0.009 (p $<$ 0.010)          & 0.651 $\pm$ 0.014 (p $=$ 0.062)          \\
ModelSoup   & 0.654 $\pm$ 0.009 (p $<$ 0.010)          & 0.618 $\pm$ 0.009 (p $<$ 0.010)          & 0.636 $\pm$ 0.008 (p $<$ 0.010)          & 0.627 $\pm$ 0.008 (p $<$ 0.010)          & 0.647 $\pm$ 0.008 (p $<$ 0.010)          & 0.637 $\pm$ 0.015 (p $=$ 0.062)          \\
FedAvg      & 0.671 $\pm$ 0.008 (p $<$ 0.010)          & 0.670 $\pm$ 0.008 (p $<$ 0.010)          & 0.675 $\pm$ 0.008 (p $<$ 0.010)          & \textbf{0.675 $\pm$ 0.008 (p $=$ 1.000)} & 0.665 $\pm$ 0.008 (p $=$ 0.051)          & \textbf{0.671 $\pm$ 0.004 (p $=$ 0.125)} \\
FedProx     & 0.668 $\pm$ 0.008 (p $<$ 0.010)          & \textbf{0.678 $\pm$ 0.008 (p $=$ 1.000)} & 0.683 $\pm$ 0.009 (p $<$ 0.010)          & 0.658 $\pm$ 0.009 (p $<$ 0.010)          & 0.661 $\pm$ 0.008 (p $<$ 0.010)          & \textbf{0.670 $\pm$ 0.010 (p $=$ 0.188)} \\
SCAFFOLD    & 0.672 $\pm$ 0.009 (p $<$ 0.010)          & 0.661 $\pm$ 0.009 (p $<$ 0.010)          & 0.671 $\pm$ 0.008 (p $<$ 0.010)          & 0.670 $\pm$ 0.009 (p $<$ 0.010)          & 0.659 $\pm$ 0.007 (p $<$ 0.010)          & 0.666 $\pm$ 0.007 (p $=$ 0.062)          \\
\end{tabular}
}

    \label{tab:strong_patch_accuracy_std_Full}
\end{table}

% Test set Low

\begin{table}[H]
    \centering
    \caption{\small Performance of the patch classifiers trained on the strongly heterogeneous setting with 2 clients and evaluated on the test set filtered on low densities, in terms of the AUC-ROC One-vs-One (mean $\pm$ covariance). Bold values indicate the best-performing models, whose differences in performance are not statistically significant (Wilcoxon Signed Rank test p-values greater than 0.1).}
    \caption*{\textbf{Strong setting (2 clients) - Patch Clfs. - Low densities AUC-ROC OvO}}
    \resizebox{\textwidth}{!}{
\begin{tabular}{lcccccc}
\toprule
Model       & Fold 1                                   & Fold 2                                   & Fold 3                                   & Fold 4                                   & Fold 5                                   & CV                                  \\
\midrule
LocalLow    & 0.889 $\pm$ 0.007 (p $<$ 0.010)          & 0.897 $\pm$ 0.007 (p $<$ 0.010)          & 0.895 $\pm$ 0.007 (p $<$ 0.010)          & 0.877 $\pm$ 0.008 (p $<$ 0.010)          & 0.896 $\pm$ 0.007 (p $<$ 0.010)          & 0.891 $\pm$ 0.009 (p $=$ 0.062)          \\
LocalHigh   & 0.877 $\pm$ 0.006 (p $<$ 0.010)          & 0.879 $\pm$ 0.008 (p $<$ 0.010)          & 0.874 $\pm$ 0.007 (p $<$ 0.010)          & 0.869 $\pm$ 0.007 (p $<$ 0.010)          & 0.884 $\pm$ 0.007 (p $<$ 0.010)          & 0.877 $\pm$ 0.006 (p $=$ 0.062)          \\
Centralized & \textbf{0.903 $\pm$ 0.007 (p $=$ 1.000)} & \textbf{0.911 $\pm$ 0.007 (p $=$ 1.000)} & \textbf{0.913 $\pm$ 0.006 (p $=$ 0.475)} & \textbf{0.902 $\pm$ 0.007 (p $=$ 1.000)} & 0.899 $\pm$ 0.008 (p $=$ 0.013)          & \textbf{0.905 $\pm$ 0.006 (p $=$ 1.000)} \\
Ensemble    & 0.873 $\pm$ 0.007 (p $<$ 0.010)          & 0.875 $\pm$ 0.009 (p $<$ 0.010)          & 0.872 $\pm$ 0.007 (p $<$ 0.010)          & 0.864 $\pm$ 0.009 (p $<$ 0.010)          & 0.881 $\pm$ 0.008 (p $<$ 0.010)          & 0.873 $\pm$ 0.006 (p $=$ 0.062)          \\
ModelSoup   & 0.885 $\pm$ 0.007 (p $<$ 0.010)          & 0.888 $\pm$ 0.007 (p $<$ 0.010)          & 0.890 $\pm$ 0.007 (p $<$ 0.010)          & 0.874 $\pm$ 0.007 (p $<$ 0.010)          & 0.900 $\pm$ 0.006 (p $<$ 0.010)          & \textbf{0.887 $\pm$ 0.009 (p $=$ 0.125)} \\
FedAvg      & 0.895 $\pm$ 0.006 (p $<$ 0.010)          & 0.906 $\pm$ 0.006 (p $<$ 0.010)          & \textbf{0.914 $\pm$ 0.006 (p $=$ 1.000)} & 0.897 $\pm$ 0.006 (p $<$ 0.010)          & 0.899 $\pm$ 0.007 (p $<$ 0.010)          & \textbf{0.902 $\pm$ 0.008 (p $=$ 0.312)} \\
FedProx     & 0.897 $\pm$ 0.006 (p $<$ 0.010)          & 0.902 $\pm$ 0.008 (p $<$ 0.010)          & 0.908 $\pm$ 0.007 (p $<$ 0.010)          & 0.889 $\pm$ 0.009 (p $<$ 0.010)          & 0.893 $\pm$ 0.007 (p $<$ 0.010)          & 0.898 $\pm$ 0.007 (p $=$ 0.062)          \\
SCAFFOLD    & 0.895 $\pm$ 0.006 (p $<$ 0.010)          & 0.901 $\pm$ 0.006 (p $<$ 0.010)          & 0.906 $\pm$ 0.007 (p $<$ 0.010)          & 0.897 $\pm$ 0.007 (p $<$ 0.010)          & \textbf{0.902 $\pm$ 0.007 (p $=$ 1.000)} & \textbf{0.900 $\pm$ 0.005 (p $=$ 0.125)} \\
\end{tabular}
}

    \label{tab:strong_patch_roc_auc_ovo_std_Low}
\end{table}

\begin{table}[H]
    \centering
    \caption{\small Performance of the patch classifiers trained on the strongly heterogeneous setting with 2 clients and evaluated on the test set filtered on low densities, in terms of the AUC-ROC One-vs-Rest (mean $\pm$ covariance). Bold values indicate the best-performing models, whose differences in performance are not statistically significant (Wilcoxon Signed Rank test p-values greater than 0.1).}
    \caption*{\textbf{Strong setting (2 clients) - Patch Clfs. - Low densities AUC-ROC OvR}}
    \resizebox{\textwidth}{!}{
\begin{tabular}{lcccccc}
\toprule
Model       & Fold 1                                   & Fold 2                                   & Fold 3                                   & Fold 4                                   & Fold 5                                   & CV                                  \\
\midrule
LocalLow    & 0.848 $\pm$ 0.011 (p $<$ 0.010)          & 0.866 $\pm$ 0.010 (p $<$ 0.010)          & 0.851 $\pm$ 0.010 (p $<$ 0.010)          & 0.824 $\pm$ 0.011 (p $<$ 0.010)          & 0.834 $\pm$ 0.010 (p $<$ 0.010)          & 0.844 $\pm$ 0.016 (p $=$ 0.062)          \\
LocalHigh   & 0.835 $\pm$ 0.010 (p $<$ 0.010)          & 0.852 $\pm$ 0.009 (p $<$ 0.010)          & 0.823 $\pm$ 0.009 (p $<$ 0.010)          & 0.831 $\pm$ 0.009 (p $<$ 0.010)          & 0.852 $\pm$ 0.008 (p $<$ 0.010)          & 0.839 $\pm$ 0.013 (p $=$ 0.062)          \\
Centralized & 0.865 $\pm$ 0.010 (p $<$ 0.010)          & \textbf{0.879 $\pm$ 0.010 (p $=$ 0.392)} & \textbf{0.872 $\pm$ 0.009 (p $=$ 1.000)} & 0.852 $\pm$ 0.011 (p $<$ 0.010)          & \textbf{0.859 $\pm$ 0.010 (p $=$ 1.000)} & \textbf{0.864 $\pm$ 0.010 (p $=$ 1.000)} \\
Ensemble    & 0.838 $\pm$ 0.010 (p $<$ 0.010)          & 0.856 $\pm$ 0.011 (p $<$ 0.010)          & 0.833 $\pm$ 0.010 (p $<$ 0.010)          & 0.830 $\pm$ 0.011 (p $<$ 0.010)          & 0.840 $\pm$ 0.011 (p $<$ 0.010)          & 0.840 $\pm$ 0.010 (p $=$ 0.062)          \\
ModelSoup   & 0.845 $\pm$ 0.009 (p $<$ 0.010)          & 0.863 $\pm$ 0.009 (p $<$ 0.010)          & 0.851 $\pm$ 0.011 (p $<$ 0.010)          & 0.818 $\pm$ 0.010 (p $<$ 0.010)          & 0.848 $\pm$ 0.010 (p $<$ 0.010)          & 0.845 $\pm$ 0.016 (p $=$ 0.062)          \\
FedAvg      & 0.853 $\pm$ 0.009 (p $<$ 0.010)          & \textbf{0.881 $\pm$ 0.008 (p $=$ 1.000)} & 0.870 $\pm$ 0.009 (p $=$ 0.075)          & 0.848 $\pm$ 0.010 (p $<$ 0.010)          & 0.852 $\pm$ 0.010 (p $<$ 0.010)          & \textbf{0.861 $\pm$ 0.014 (p $=$ 0.312)} \\
FedProx     & \textbf{0.871 $\pm$ 0.008 (p $=$ 1.000)} & 0.870 $\pm$ 0.010 (p $<$ 0.010)          & 0.869 $\pm$ 0.010 (p $=$ 0.014)          & \textbf{0.861 $\pm$ 0.011 (p $=$ 1.000)} & 0.853 $\pm$ 0.010 (p $<$ 0.010)          & \textbf{0.864 $\pm$ 0.007 (p $=$ 1.000)} \\
SCAFFOLD    & \textbf{0.870 $\pm$ 0.009 (p $=$ 0.786)} & 0.872 $\pm$ 0.008 (p $<$ 0.010)          & \textbf{0.872 $\pm$ 0.009 (p $=$ 1.000)} & 0.847 $\pm$ 0.010 (p $<$ 0.010)          & \textbf{0.857 $\pm$ 0.010 (p $=$ 0.134)} & \textbf{0.863 $\pm$ 0.011 (p $=$ 0.812)} \\
\end{tabular}
}

    \label{tab:strong_patch_roc_auc_AUC-ROC OvR_std_Low}
\end{table}

\begin{table}[H]
    \centering
    \caption{\small Performance of the patch classifiers trained on the strongly heterogeneous setting with 2 clients and evaluated on the test set filtered on low densities, in terms of the accuracy (mean $\pm$ covariance). Bold values indicate the best-performing models, whose differences in performance are not statistically significant (Wilcoxon Signed Rank test p-values greater than 0.1).}
    \caption*{\textbf{Strong setting (2 clients) - Patch Clfs. - Low densities Accuracy}}
    \resizebox{\textwidth}{!}{
\begin{tabular}{lcccccc}
\toprule
Model       & Fold 1                                   & Fold 2                                   & Fold 3                                   & Fold 4                                   & Fold 5                                   & CV                                  \\
\midrule
LocalLow    & 0.669 $\pm$ 0.014 (p $<$ 0.010)          & 0.660 $\pm$ 0.015 (p $<$ 0.010)          & 0.688 $\pm$ 0.015 (p $<$ 0.010)          & 0.653 $\pm$ 0.014 (p $<$ 0.010)          & 0.675 $\pm$ 0.015 (p $<$ 0.010)          & 0.669 $\pm$ 0.014 (p $=$ 0.062)          \\
LocalHigh   & 0.667 $\pm$ 0.015 (p $<$ 0.010)          & 0.625 $\pm$ 0.015 (p $<$ 0.010)          & 0.615 $\pm$ 0.014 (p $<$ 0.010)          & 0.610 $\pm$ 0.016 (p $<$ 0.010)          & 0.644 $\pm$ 0.015 (p $<$ 0.010)          & 0.632 $\pm$ 0.024 (p $=$ 0.062)          \\
Centralized & \textbf{0.685 $\pm$ 0.014 (p $=$ 1.000)} & \textbf{0.706 $\pm$ 0.015 (p $=$ 1.000)} & \textbf{0.715 $\pm$ 0.013 (p $=$ 0.107)} & \textbf{0.696 $\pm$ 0.015 (p $=$ 1.000)} & 0.678 $\pm$ 0.016 (p $<$ 0.010)          & \textbf{0.695 $\pm$ 0.015 (p $=$ 1.000)} \\
Ensemble    & 0.674 $\pm$ 0.013 (p $<$ 0.010)          & 0.642 $\pm$ 0.016 (p $<$ 0.010)          & 0.665 $\pm$ 0.015 (p $<$ 0.010)          & 0.645 $\pm$ 0.016 (p $<$ 0.010)          & 0.674 $\pm$ 0.014 (p $<$ 0.010)          & 0.660 $\pm$ 0.016 (p $=$ 0.062)          \\
ModelSoup   & 0.664 $\pm$ 0.014 (p $<$ 0.010)          & 0.623 $\pm$ 0.017 (p $<$ 0.010)          & 0.645 $\pm$ 0.017 (p $<$ 0.010)          & 0.632 $\pm$ 0.015 (p $<$ 0.010)          & 0.657 $\pm$ 0.013 (p $<$ 0.010)          & 0.644 $\pm$ 0.016 (p $=$ 0.062)          \\
FedAvg      & 0.674 $\pm$ 0.015 (p $<$ 0.010)          & 0.686 $\pm$ 0.013 (p $<$ 0.010)          & 0.708 $\pm$ 0.014 (p $<$ 0.010)          & \textbf{0.696 $\pm$ 0.013 (p $=$ 1.000)} & \textbf{0.687 $\pm$ 0.015 (p $=$ 1.000)} & \textbf{0.690 $\pm$ 0.013 (p $=$ 0.438)} \\
FedProx     & 0.677 $\pm$ 0.013 (p $<$ 0.010)          & 0.696 $\pm$ 0.015 (p $<$ 0.010)          & \textbf{0.718 $\pm$ 0.014 (p $=$ 1.000)} & 0.676 $\pm$ 0.016 (p $<$ 0.010)          & 0.676 $\pm$ 0.014 (p $<$ 0.010)          & \textbf{0.689 $\pm$ 0.018 (p $=$ 0.312)} \\
SCAFFOLD    & 0.674 $\pm$ 0.015 (p $<$ 0.010)          & 0.675 $\pm$ 0.013 (p $<$ 0.010)          & 0.706 $\pm$ 0.012 (p $<$ 0.010)          & 0.689 $\pm$ 0.015 (p $<$ 0.010)          & 0.680 $\pm$ 0.014 (p $<$ 0.010)          & \textbf{0.686 $\pm$ 0.013 (p $=$ 0.188)} \\
\end{tabular}
}

    \label{tab:strong_patch_accuracy_std_Low}
\end{table}

 % Test set High

\begin{table}[H]
    \centering
    \caption{\small Performance of the patch classifiers trained on the strongly heterogeneous setting with 2 clients and evaluated on the test set filtered on high densities, in terms of the AUC-ROC One-vs-One (mean $\pm$ covariance). Bold values indicate the best-performing models, whose differences in performance are not statistically significant (Wilcoxon Signed Rank test p-values greater than 0.1).}
    \caption*{\textbf{Strong setting (2 clients) - Patch Clfs. - High densities AUC-ROC OvO}}
    \resizebox{\textwidth}{!}{
\begin{tabular}{lcccccc}
\toprule
Model       & Fold 1                                   & Fold 2                                   & Fold 3                                   & Fold 4                                   & Fold 5                                   & CV                                  \\
\midrule
LocalLow    & 0.898 $\pm$ 0.005 (p $<$ 0.010)          & 0.905 $\pm$ 0.005 (p $<$ 0.010)          & 0.907 $\pm$ 0.005 (p $<$ 0.010)          & 0.891 $\pm$ 0.006 (p $<$ 0.010)          & 0.901 $\pm$ 0.005 (p $<$ 0.010)          & 0.901 $\pm$ 0.006 (p $=$ 0.062)          \\
LocalHigh   & 0.903 $\pm$ 0.005 (p $<$ 0.010)          & 0.906 $\pm$ 0.005 (p $<$ 0.010)          & 0.908 $\pm$ 0.005 (p $<$ 0.010)          & 0.909 $\pm$ 0.005 (p $<$ 0.010)          & 0.904 $\pm$ 0.005 (p $<$ 0.010)          & 0.906 $\pm$ 0.002 (p $=$ 0.062)          \\
Centralized & \textbf{0.920 $\pm$ 0.005 (p $=$ 1.000)} & \textbf{0.915 $\pm$ 0.005 (p $=$ 1.000)} & \textbf{0.921 $\pm$ 0.005 (p $=$ 1.000)} & \textbf{0.923 $\pm$ 0.005 (p $=$ 1.000)} & \textbf{0.914 $\pm$ 0.005 (p $=$ 1.000)} & \textbf{0.919 $\pm$ 0.004 (p $=$ 1.000)} \\
Ensemble    & 0.899 $\pm$ 0.006 (p $<$ 0.010)          & 0.900 $\pm$ 0.005 (p $<$ 0.010)          & 0.907 $\pm$ 0.004 (p $<$ 0.010)          & 0.899 $\pm$ 0.005 (p $<$ 0.010)          & 0.901 $\pm$ 0.005 (p $<$ 0.010)          & 0.902 $\pm$ 0.003 (p $=$ 0.062)          \\
ModelSoup   & 0.909 $\pm$ 0.005 (p $<$ 0.010)          & 0.908 $\pm$ 0.005 (p $<$ 0.010)          & 0.912 $\pm$ 0.005 (p $<$ 0.010)          & 0.906 $\pm$ 0.005 (p $<$ 0.010)          & \textbf{0.914 $\pm$ 0.004 (p $=$ 1.000)} & 0.909 $\pm$ 0.003 (p $=$ 0.062)          \\
FedAvg      & 0.918 $\pm$ 0.005 (p $=$ 0.014)          & \textbf{0.915 $\pm$ 0.005 (p $=$ 1.000)} & 0.917 $\pm$ 0.005 (p $<$ 0.010)          & 0.919 $\pm$ 0.005 (p $<$ 0.010)          & 0.911 $\pm$ 0.005 (p $<$ 0.010)          & 0.916 $\pm$ 0.003 (p $=$ 0.062)          \\
FedProx     & 0.918 $\pm$ 0.005 (p $=$ 0.037)          & 0.913 $\pm$ 0.005 (p $<$ 0.010)          & 0.914 $\pm$ 0.005 (p $<$ 0.010)          & 0.915 $\pm$ 0.005 (p $<$ 0.010)          & 0.903 $\pm$ 0.005 (p $<$ 0.010)          & 0.912 $\pm$ 0.005 (p $=$ 0.062)          \\
SCAFFOLD    & \textbf{0.919 $\pm$ 0.005 (p $=$ 0.466)} & 0.912 $\pm$ 0.005 (p $<$ 0.010)          & 0.912 $\pm$ 0.005 (p $<$ 0.010)          & 0.917 $\pm$ 0.005 (p $<$ 0.010)          & \textbf{0.914 $\pm$ 0.004 (p $=$ 1.000)} & \textbf{0.915 $\pm$ 0.003 (p $=$ 0.125)} \\
\end{tabular}
}

    \label{tab:strong_patch_roc_auc_ovo_std_High}
\end{table}

\begin{table}[H]
    \centering
    \caption{\small Performance of the patch classifiers trained on the strongly heterogeneous setting with 2 clients and evaluated on the test set filtered on high densities, in terms of the AUC-ROC One-vs-Rest (mean $\pm$ covariance). Bold values indicate the best-performing models, whose differences in performance are not statistically significant (Wilcoxon Signed Rank test p-values greater than 0.1).}
    \caption*{\textbf{Strong setting (2 clients) - Patch Clfs. - High densities AUC-ROC OvR}}
    \resizebox{\textwidth}{!}{
\begin{tabular}{lcccccc}
\toprule
Model       & Fold 1                                   & Fold 2                                   & Fold 3                                   & Fold 4                                   & Fold 5                                   & CV                                  \\
\midrule
LocalLow    & 0.848 $\pm$ 0.006 (p $<$ 0.010)          & 0.859 $\pm$ 0.006 (p $<$ 0.010)          & 0.852 $\pm$ 0.007 (p $<$ 0.010)          & 0.833 $\pm$ 0.007 (p $<$ 0.010)          & 0.839 $\pm$ 0.006 (p $<$ 0.010)          & 0.847 $\pm$ 0.010 (p $=$ 0.062)          \\
LocalHigh   & 0.855 $\pm$ 0.007 (p $<$ 0.010)          & 0.858 $\pm$ 0.006 (p $<$ 0.010)          & 0.853 $\pm$ 0.007 (p $<$ 0.010)          & 0.862 $\pm$ 0.007 (p $<$ 0.010)          & 0.858 $\pm$ 0.007 (p $<$ 0.010)          & 0.857 $\pm$ 0.003 (p $=$ 0.062)          \\
Centralized & 0.876 $\pm$ 0.006 (p $<$ 0.010)          & 0.871 $\pm$ 0.007 (p $<$ 0.010)          & 0.874 $\pm$ 0.007 (p $<$ 0.010)          & \textbf{0.880 $\pm$ 0.007 (p $=$ 0.301)} & 0.870 $\pm$ 0.007 (p $<$ 0.010)          & \textbf{0.874 $\pm$ 0.004 (p $=$ 0.125)} \\
Ensemble    & 0.856 $\pm$ 0.008 (p $<$ 0.010)          & 0.860 $\pm$ 0.007 (p $<$ 0.010)          & 0.861 $\pm$ 0.006 (p $<$ 0.010)          & 0.857 $\pm$ 0.007 (p $<$ 0.010)          & 0.854 $\pm$ 0.007 (p $<$ 0.010)          & 0.858 $\pm$ 0.003 (p $=$ 0.062)          \\
ModelSoup   & 0.860 $\pm$ 0.007 (p $<$ 0.010)          & 0.862 $\pm$ 0.007 (p $<$ 0.010)          & 0.861 $\pm$ 0.007 (p $<$ 0.010)          & 0.856 $\pm$ 0.007 (p $<$ 0.010)          & 0.856 $\pm$ 0.007 (p $<$ 0.010)          & 0.859 $\pm$ 0.003 (p $=$ 0.062)          \\
FedAvg      & 0.876 $\pm$ 0.006 (p $<$ 0.010)          & \textbf{0.877 $\pm$ 0.006 (p $=$ 1.000)} & 0.873 $\pm$ 0.006 (p $<$ 0.010)          & 0.880 $\pm$ 0.006 (p $=$ 0.061)          & \textbf{0.876 $\pm$ 0.005 (p $=$ 1.000)} & \textbf{0.877 $\pm$ 0.002 (p $=$ 1.000)} \\
FedProx     & \textbf{0.881 $\pm$ 0.006 (p $=$ 1.000)} & \textbf{0.877 $\pm$ 0.006 (p $=$ 1.000)} & \textbf{0.879 $\pm$ 0.006 (p $=$ 1.000)} & \textbf{0.880 $\pm$ 0.006 (p $=$ 0.137)} & 0.872 $\pm$ 0.006 (p $<$ 0.010)          & \textbf{0.877 $\pm$ 0.003 (p $=$ 1.000)} \\
SCAFFOLD    & \textbf{0.881 $\pm$ 0.006 (p $=$ 1.000)} & 0.874 $\pm$ 0.006 (p $<$ 0.010)          & 0.870 $\pm$ 0.007 (p $<$ 0.010)          & \textbf{0.882 $\pm$ 0.006 (p $=$ 1.000)} & \textbf{0.876 $\pm$ 0.006 (p $=$ 1.000)} & \textbf{0.876 $\pm$ 0.005 (p $=$ 1.000)} \\
\end{tabular}
}

    \label{tab:strong_patch_roc_auc_AUC-ROC OvR_std_High}
\end{table}

\begin{table}[H]
    \centering
    \caption{\small Performance of the patch classifiers trained on the strongly heterogeneous setting with 2 clients and evaluated on the test set filtered on high densities, in terms of the accuracy (mean $\pm$ covariance). Bold values indicate the best-performing models, whose differences in performance are not statistically significant (Wilcoxon Signed Rank test p-values greater than 0.1).}
    \caption*{\textbf{Strong setting (2 clients) - Patch Clfs. - High densities Accuracy}}
    \resizebox{\textwidth}{!}{
\begin{tabular}{lcccccc}
\toprule
Model       & Fold 1                                   & Fold 2                                   & Fold 3                                   & Fold 4                                   & Fold 5                                   & CV                                  \\
\midrule
LocalLow    & 0.620 $\pm$ 0.010 (p $<$ 0.010)          & 0.626 $\pm$ 0.011 (p $<$ 0.010)          & 0.649 $\pm$ 0.011 (p $<$ 0.010)          & 0.588 $\pm$ 0.011 (p $<$ 0.010)          & 0.631 $\pm$ 0.009 (p $<$ 0.010)          & 0.623 $\pm$ 0.023 (p $=$ 0.062)          \\
LocalHigh   & 0.648 $\pm$ 0.011 (p $<$ 0.010)          & 0.593 $\pm$ 0.012 (p $<$ 0.010)          & 0.628 $\pm$ 0.011 (p $<$ 0.010)          & 0.647 $\pm$ 0.011 (p $<$ 0.010)          & 0.609 $\pm$ 0.011 (p $<$ 0.010)          & 0.625 $\pm$ 0.023 (p $=$ 0.062)          \\
Centralized & \textbf{0.673 $\pm$ 0.011 (p $=$ 1.000)} & 0.655 $\pm$ 0.010 (p $<$ 0.010)          & \textbf{0.675 $\pm$ 0.011 (p $=$ 1.000)} & \textbf{0.666 $\pm$ 0.010 (p $=$ 0.354)} & \textbf{0.662 $\pm$ 0.011 (p $=$ 1.000)} & \textbf{0.666 $\pm$ 0.008 (p $=$ 1.000)} \\
Ensemble    & 0.661 $\pm$ 0.011 (p $<$ 0.010)          & 0.626 $\pm$ 0.010 (p $<$ 0.010)          & 0.657 $\pm$ 0.009 (p $<$ 0.010)          & 0.641 $\pm$ 0.012 (p $<$ 0.010)          & 0.646 $\pm$ 0.009 (p $<$ 0.010)          & 0.647 $\pm$ 0.014 (p $=$ 0.062)          \\
ModelSoup   & 0.650 $\pm$ 0.011 (p $<$ 0.010)          & 0.615 $\pm$ 0.012 (p $<$ 0.010)          & 0.632 $\pm$ 0.010 (p $<$ 0.010)          & 0.627 $\pm$ 0.010 (p $<$ 0.010)          & 0.645 $\pm$ 0.012 (p $<$ 0.010)          & 0.633 $\pm$ 0.014 (p $=$ 0.062)          \\
FedAvg      & 0.670 $\pm$ 0.010 (p $=$ 0.024)          & 0.661 $\pm$ 0.010 (p $<$ 0.010)          & 0.653 $\pm$ 0.010 (p $<$ 0.010)          & \textbf{0.668 $\pm$ 0.010 (p $=$ 1.000)} & 0.654 $\pm$ 0.010 (p $<$ 0.010)          & \textbf{0.661 $\pm$ 0.007 (p $=$ 0.438)} \\
FedProx     & 0.664 $\pm$ 0.011 (p $<$ 0.010)          & \textbf{0.670 $\pm$ 0.011 (p $=$ 1.000)} & 0.665 $\pm$ 0.010 (p $<$ 0.010)          & 0.650 $\pm$ 0.011 (p $<$ 0.010)          & 0.655 $\pm$ 0.010 (p $<$ 0.010)          & \textbf{0.660 $\pm$ 0.008 (p $=$ 0.438)} \\
SCAFFOLD    & \textbf{0.672 $\pm$ 0.010 (p $=$ 0.301)} & 0.650 $\pm$ 0.011 (p $<$ 0.010)          & 0.653 $\pm$ 0.011 (p $<$ 0.010)          & 0.659 $\pm$ 0.011 (p $<$ 0.010)          & 0.648 $\pm$ 0.010 (p $<$ 0.010)          & 0.656 $\pm$ 0.009 (p $=$ 0.062)          \\
\end{tabular}
}

    \label{tab:strong_patch_accuracy_std_High}
\end{table}

\paragraph{Whole image classifier}

% Test set Low

\begin{table}[H]
\centering
\caption{\small Performance of the whole image classifiers trained on the strongly heterogeneous setting with 2 clients setting and evaluated on the test set filtered on low densities, in terms of the AUC-ROC  (mean $\pm$ covariance). Bold values indicate the best-performing models, whose differences in performance are not statistically significant (Wilcoxon Signed Rank test p-values greater than 0.1).}
\caption*{\textbf{Strong setting (2 clients) - Whole Image Clfs. - Low densities AUC-ROC}}
\resizebox{\textwidth}{!}{
\begin{tabular}{lcccccc}
\toprule
Model       & Fold 1                                   & Fold 2                                   & Fold 3                                   & Fold 4                                   & Fold 5                                   & CV                                  \\
\midrule
LocalLow    & 0.803 $\pm$ 0.036 (p $=$ 0.050)          & 0.801 $\pm$ 0.035 (p $<$ 0.010)          & 0.768 $\pm$ 0.043 (p $<$ 0.010)          & 0.781 $\pm$ 0.038 (p $<$ 0.010)          & 0.733 $\pm$ 0.043 (p $<$ 0.010)          & \textbf{0.783 $\pm$ 0.028 (p $=$ 0.188)} \\
LocalHigh   & 0.715 $\pm$ 0.051 (p $<$ 0.010)          & 0.735 $\pm$ 0.049 (p $<$ 0.010)          & 0.770 $\pm$ 0.048 (p $<$ 0.010)          & 0.678 $\pm$ 0.063 (p $<$ 0.010)          & 0.767 $\pm$ 0.050 (p $<$ 0.010)          & 0.737 $\pm$ 0.034 (p $=$ 0.062)          \\
Centralized & 0.781 $\pm$ 0.035 (p $<$ 0.010)          & 0.749 $\pm$ 0.035 (p $<$ 0.010)          & 0.780 $\pm$ 0.038 (p $<$ 0.010)          & 0.753 $\pm$ 0.038 (p $<$ 0.010)          & 0.734 $\pm$ 0.051 (p $<$ 0.010)          & 0.762 $\pm$ 0.018 (p $=$ 0.062)          \\
Ensemble    & \textbf{0.814 $\pm$ 0.036 (p $=$ 1.000)} & \textbf{0.825 $\pm$ 0.036 (p $=$ 0.989)} & \textbf{0.800 $\pm$ 0.038 (p $=$ 0.282)} & 0.782 $\pm$ 0.039 (p $=$ 0.026)          & 0.792 $\pm$ 0.040 (p $<$ 0.010)          & \textbf{0.799 $\pm$ 0.016 (p $=$ 1.000)} \\
ModelSoup   & 0.649 $\pm$ 0.075 (p $<$ 0.010)          & 0.788 $\pm$ 0.043 (p $<$ 0.010)          & \textbf{0.805 $\pm$ 0.047 (p $=$ 1.000)} & 0.754 $\pm$ 0.056 (p $<$ 0.010)          & 0.784 $\pm$ 0.050 (p $<$ 0.010)          & \textbf{0.754 $\pm$ 0.064 (p $=$ 0.125)} \\
FedAvg      & 0.805 $\pm$ 0.037 (p $=$ 0.058)          & \textbf{0.826 $\pm$ 0.036 (p $=$ 1.000)} & 0.790 $\pm$ 0.038 (p $=$ 0.020)          & \textbf{0.787 $\pm$ 0.040 (p $=$ 0.180)} & 0.765 $\pm$ 0.040 (p $<$ 0.010)          & \textbf{0.795 $\pm$ 0.022 (p $=$ 0.625)} \\
FedProx     & 0.768 $\pm$ 0.048 (p $<$ 0.010)          & 0.797 $\pm$ 0.040 (p $<$ 0.010)          & \textbf{0.796 $\pm$ 0.042 (p $=$ 0.126)} & \textbf{0.796 $\pm$ 0.038 (p $=$ 1.000)} & 0.757 $\pm$ 0.041 (p $<$ 0.010)          & \textbf{0.785 $\pm$ 0.018 (p $=$ 0.312)} \\
SCAFFOLD    & \textbf{0.813 $\pm$ 0.047 (p $=$ 0.929)} & 0.775 $\pm$ 0.042 (p $<$ 0.010)          & 0.786 $\pm$ 0.042 (p $<$ 0.010)          & 0.760 $\pm$ 0.045 (p $<$ 0.010)          & \textbf{0.819 $\pm$ 0.034 (p $=$ 1.000)} & \textbf{0.791 $\pm$ 0.024 (p $=$ 0.625)} \\
\end{tabular}
}

\label{tab:strong_2_clients_roc_auc_std_Low}
\end{table}

% Test set High

\begin{table}[H]
\centering
\caption{\small Performance of the whole image classifiers trained on the strongly heterogeneous setting with 2 clients setting and evaluated on the test set filtered on high densities, in terms of the AUC-ROC  (mean $\pm$ covariance). Bold values indicate the best-performing models, whose differences in performance are not statistically significant (Wilcoxon Signed Rank test p-values greater than 0.1).}
\caption*{\textbf{Strong setting (2 clients) - Whole Image Clfs. - High densities perf.}}
\resizebox{\textwidth}{!}{
\begin{tabular}{lcccccc}
\toprule
Model       & Fold 1                                   & Fold 2                                   & Fold 3                                   & Fold 4                                   & Fold 5                                   & CV                                  \\
\midrule
LocalLow    & 0.710 $\pm$ 0.040 (p $<$ 0.010)          & \textbf{0.759 $\pm$ 0.030 (p $=$ 1.000)} & 0.751 $\pm$ 0.036 (p $<$ 0.010)          & 0.741 $\pm$ 0.039 (p $=$ 0.014)          & 0.745 $\pm$ 0.036 (p $<$ 0.010)          & 0.743 $\pm$ 0.019 (p $=$ 0.062)          \\
LocalHigh   & 0.729 $\pm$ 0.038 (p $<$ 0.010)          & 0.718 $\pm$ 0.035 (p $<$ 0.010)          & 0.730 $\pm$ 0.042 (p $<$ 0.010)          & 0.733 $\pm$ 0.041 (p $<$ 0.010)          & 0.737 $\pm$ 0.042 (p $<$ 0.010)          & 0.731 $\pm$ 0.005 (p $=$ 0.062)          \\
Centralized & 0.751 $\pm$ 0.041 (p $<$ 0.010)          & 0.724 $\pm$ 0.043 (p $<$ 0.010)          & 0.751 $\pm$ 0.039 (p $<$ 0.010)          & 0.743 $\pm$ 0.047 (p $=$ 0.095)          & \textbf{0.763 $\pm$ 0.037 (p $=$ 0.182)} & \textbf{0.750 $\pm$ 0.016 (p $=$ 0.625)} \\
Ensemble    & 0.733 $\pm$ 0.043 (p $<$ 0.010)          & \textbf{0.759 $\pm$ 0.033 (p $=$ 1.000)} & \textbf{0.764 $\pm$ 0.034 (p $=$ 0.578)} & \textbf{0.757 $\pm$ 0.039 (p $=$ 1.000)} & 0.754 $\pm$ 0.032 (p $<$ 0.010)          & \textbf{0.759 $\pm$ 0.013 (p $=$ 1.000)} \\
ModelSoup   & 0.707 $\pm$ 0.045 (p $<$ 0.010)          & 0.746 $\pm$ 0.039 (p $<$ 0.010)          & \textbf{0.767 $\pm$ 0.035 (p $=$ 1.000)} & 0.742 $\pm$ 0.037 (p $=$ 0.018)          & \textbf{0.771 $\pm$ 0.042 (p $=$ 1.000)} & \textbf{0.749 $\pm$ 0.027 (p $=$ 0.312)} \\
FedAvg      & \textbf{0.770 $\pm$ 0.036 (p $=$ 1.000)} & 0.746 $\pm$ 0.037 (p $<$ 0.010)          & \textbf{0.763 $\pm$ 0.044 (p $=$ 0.384)} & 0.726 $\pm$ 0.042 (p $<$ 0.010)          & 0.734 $\pm$ 0.044 (p $<$ 0.010)          & \textbf{0.751 $\pm$ 0.018 (p $=$ 0.625)} \\
FedProx     & 0.756 $\pm$ 0.042 (p $=$ 0.011)          & 0.709 $\pm$ 0.052 (p $<$ 0.010)          & 0.752 $\pm$ 0.039 (p $<$ 0.010)          & 0.731 $\pm$ 0.052 (p $<$ 0.010)          & 0.746 $\pm$ 0.045 (p $<$ 0.010)          & \textbf{0.737 $\pm$ 0.019 (p $=$ 0.312)} \\
SCAFFOLD    & 0.757 $\pm$ 0.042 (p $=$ 0.015)          & \textbf{0.759 $\pm$ 0.039 (p $=$ 1.000)} & 0.751 $\pm$ 0.041 (p $<$ 0.010)          & 0.736 $\pm$ 0.053 (p $<$ 0.010)          & \textbf{0.765 $\pm$ 0.039 (p $=$ 0.276)} & \textbf{0.751 $\pm$ 0.013 (p $=$ 0.625)} \\
\end{tabular}
}

\label{tab:strong_2_clients_roc_auc_std_High}
\end{table}

% \disclosures 
\subsection*{Disclosures}
The authors declare that there are no financial interests, commercial affiliations, or other potential conflicts of interest that could have influenced the objectivity of this research or the writing of this paper, besides the ones related to the affiliations of the authors.

\subsection*{Code, Data, and Materials Availability}
All experiments were conducted using PyTorch and the NVIDIA FLARE \cite{Roth_NVIDIA_FLARE_Federated_2023}  framework for training Federated Learning models in realistic settings. The training data is not publicly available.

\subsection*{Acknowledgments}

This work was partially funded by Association Nationale de la Recherche et de la Technologie (ANRT) under CIFRE grant number 2021/0422. In addition, the authors declare having used ChatGPT to assist with grammar and style corrections when preparing this submission. All content was subsequently reviewed and edited by the authors, who take full responsibility for the final published version.

%%%%% References %%%%%

\bibliography{bib}   % bibliography data in report.bib

@article{wood2023unified,
  title={A unified theory of diversity in ensemble learning},
  author={Wood, Danny and Mu, Tingting and Webb, Andrew M and Reeve, Henry WJ and Lujan, Mikel and Brown, Gavin},
  journal={Journal of machine learning research},
  volume={24},
  number={359},
  pages={1--49},
  year={2023}
}

@phdthesis{geurts2002contributions,
  title={Contributions to decision tree induction: bias/variance tradeoff and time series classification},
  author={Geurts, Pierre},
  year={2002},
  school={Universite de Liege (Belgium)}
}

@book{goodfellow2016deep,
  title={Deep learning},
  author={Goodfellow, Ian and Bengio, Yoshua and Courville, Aaron and Bengio, Yoshua},
  volume={1},
  number={2},
  year={2016},
  publisher={MIT press Cambridge}
}

@inproceedings{izmailov2018averaging,
  title={Averaging weights leads to wider optima and better generalization},
  author={Izmailov, Pavel and Podoprikhin, Dmitrii and Garipov, Timur and Vetrov, Dmitry and Wilson, Andrew Gordon},
  booktitle={34th Conference on Uncertainty in Artificial Intelligence 2018, UAI 2018},
  pages={876--885},
  year={2018},
  organization={Association For Uncertainty in Artificial Intelligence (AUAI)}
}

@article{garipov2018loss,
  title={Loss surfaces, mode connectivity, and fast ensembling of dnns},
  author={Garipov, Timur and Izmailov, Pavel and Podoprikhin, Dmitrii and Vetrov, Dmitry P and Wilson, Andrew G},
  journal={Advances in neural information processing systems},
  volume={31},
  year={2018}
}

@article{kerlikowske2023impact,
  title={Impact of BMI on Prevalence of Dense Breasts by Race and Ethnicity},
  author={Karla Kerlikowske and Michael C. S. Bissell and Brian L. Sprague and Jeffrey A. Tice and Katherine Y. Tossas and Erin J. A. Bowles and Thao-Quyen H. Ho and Theresa H. M. Keegan and Diana L. Miglioretti},
  journal={Cancer Epidemiology, Biomarkers and Prevention},
  volume={32},
  number={11},
  pages={1524--1530},
  year={2023},
  publisher={American Association for Cancer Research},
  doi={10.1158/1055-9965.EPI-23-0049}
}

@article{Nazari2018,
  title={An overview of mammographic density and its association with breast cancer},
  author={Nazari, Shayan Shaghayeq and Mukherjee, Pinku},
  journal={Breast Cancer},
  publisher={Springer},
  volume={25},
  number={3},
  pages={259--267},
  year={2018},
  month={May},
  doi={10.1007/s12282-018-0857-5},
  pmid={29651637},
  pmcid={PMC5906528}
}

@article{vonEulerChelpin2019,
  title={Sensitivity of screening mammography by density and texture: a cohort study from a population-based screening program in Denmark},
  author={von Euler-Chelpin, My and Lillholm, Martin and Vejborg, Inge and et al.},
  journal={Breast Cancer Research},
  volume={21},
  number={1},
  pages={111},
  year={2019},
  doi={10.1186/s13058-019-1203-3},
  url={https://doi.org/10.1186/s13058-019-1203-3}
}

@article{vanderWaal2017,
  title={Breast cancer screening effect across breast density strata: A case-control study},
  author={van der Waal, Douwe and Ripping, Tjalling M. and Verbeek, André L. M. and Broeders, Mireille J. M.},
  journal={International Journal of Cancer},
  volume={140},
  number={1},
  pages={41--49},
  year={2017},
  month={Jan},
  doi={10.1002/ijc.30430},
  pmid={27632020}
}

@article{Suh2020,
  title={Automated Breast Cancer Detection in Digital Mammograms of Various Densities via Deep Learning},
  author={Suh, Yong Joon and Jung, Jaewon and Cho, Bum-Joo},
  journal={Journal of Personalized Medicine},
  volume={10},
  number={4},
  pages={211},
  year={2020},
  doi={10.3390/jpm10040211},
  url={https://doi.org/10.3390/jpm10040211}
}

@article{Kim2023,
  title={Deep Learning Analysis of Mammography for Breast Cancer Risk Prediction in Asian Women},
  author={Kim, Hae-Won and Lim, Joon-Soo and Kim, Hae-Young and Lim, Youn-Kyung and Seo, Byung-Keun and Bae, Min-Seok},
  journal={Diagnostics},
  volume={13},
  number={13},
  pages={2247},
  year={2023},
  month={Jul},
  doi={10.3390/diagnostics13132247},
  pmid={37443642},
  pmcid={PMC10340460}
}

@article{Brem2005,
  title={Impact of Breast Density on Computer-Aided Detection for Breast Cancer},
  author={Brem, Rachel F. and Hoffmeister, Jeffrey W. and Rapelyea, Jocelyn A. and Zisman, Gilat and Mohtashemi, Kevin and Jindal, Guarav and DiSimio, Martin P. and Rogers, Steven K.},
  journal={American Journal of Roentgenology},
  volume={184},
  number={2},
  pages={404--409},
  year={2005},
  doi={10.2214/ajr.184.2.01840439},
  url={https://doi.org/10.2214/ajr.184.2.01840439}
}

@inproceedings{Dustler2020,
  author={Dustler, Magnus and Dahlblom, Victor and Tingberg, Anders and Zackrisson, Sophia},
  title={The effect of breast density on the performance of deep learning-based breast cancer detection methods for mammography},
  booktitle={Proceedings of SPIE 11513, 15th International Workshop on Breast Imaging (IWBI 2020)},
  volume={11513},
  pages={1151324},
  year={2020},
  month={May},
  publisher={SPIE},
  doi={10.1117/12.2564328}
}

@article{RiveiraMartin2023,
  title={Multi-vendor robustness analysis of a commercial artificial intelligence system for breast cancer detection},
  author={Riveira-Martin, Marta and Rodríguez-Ruiz, Adriana and Martí, Raúl and Chevalier, Max},
  journal={Journal of Medical Imaging},
  volume={10},
  number={5},
  pages={051807},
  year={2023},
  month={Sep},
  doi={10.1117/1.JMI.10.5.051807},
  eprint={2023 Apr 18},
  pmid={37082509},
  pmcid={PMC10111789}
}

@article{Yala2021,
  title={Toward robust mammography-based models for breast cancer risk},
  author={Yala, Adam and others},
  journal={Science Translational Medicine},
  volume={13},
  number={580},
  pages={eaba4373},
  year={2021},
  doi={10.1126/scitranslmed.aba4373}
}

@article{Omoleye2023,
  title={External Evaluation of a Mammography-based Deep Learning Model for Predicting Breast Cancer in an Ethnically Diverse Population},
  author={Omoleye, Olamide J. and Woodard, Alice E. and Howard, Faith M. and Zhao, Fei and Yoshimatsu, Takashi F. and Zheng, Ying and Pearson, Andrew T. and Levental, Matthew and Aribisala, Bola S. and Kulkarni, Kranti and Karczmar, Gregory S. and Olopade, Olufunmilayo I. and Abe, Hiroshi and Huo, Deling},
  journal={Radiology: Artificial Intelligence},
  volume={5},
  number={6},
  pages={e220299},
  year={2023},
  month={Jul},
  doi={10.1148/ryai.220299},
  pmid={38074785},
  pmcid={PMC10698602}
}

@article{shen2019deep,
  author    = {Shen, Li and Margolies, Laurie R. and Rothstein, Joseph H. and others},
  title     = {Deep Learning to Improve Breast Cancer Detection on Screening Mammography},
  journal   = {Scientific Reports},
  volume    = {9},
  number    = {1},
  pages     = {12495},
  year      = {2019},
  doi       = {10.1038/s41598-019-48995-4},
  url       = {https://doi.org/10.1038/s41598-019-48995-4}
}

@inproceedings{terrassin2024weakly,
  title={Weakly-supervised end-to-end framework for pixel-wise description of micro-calcifications in full-resolution mammograms},
  author={Terrassin, Paul and Tardy, Mickael and Lauzeral, Nathan and Normand, Nicolas},
  booktitle={2024 IEEE 21st International Symposium on Biomedical Imaging (ISBI)},
  year={2024},
  organization={IEEE}
}

@article{quintana2023mammo,
    author = {Quintana, Gonzalo Iñaki and Li, Zhijin and Vancamberg, Laurence and Mougeot, Mathilde and Desolneux, Agnès and Muller, Serge},
    title = {Exploiting patch sizes and resolutions for multi-scale Deep Learning in mammography image classification},
    journal = {Bioengineering},
    volume = {10},
    number = {5},
    pages = {534},
    year = {2023},
    publisher = {MDPI},
    doi = {10.3390/bioengineering10050534},
    url = {https://doi.org/10.3390/bioengineering10050534}
}

@article{petrini_2021,
  title={Breast cancer diagnosis in two-view mammography using end-to-end trained efficientnet-based convolutional network},
  author={Petrini, Daniel GP and Shimizu, Carlos and Roela, Rosimeire A and Valente, Gabriel Vansuita and Folgueira, Maria Aparecida Azevedo Koike and Kim, Hae Yong},
  journal={Ieee access},
  volume={10},
  pages={77723--77731},
  year={2022},
  publisher={IEEE}
}

@inproceedings{patch_clsf_sliding_wdw,
  title={Deep multi-instance networks with sparse label assignment for whole mammogram classification},
  author={Zhu, Wentao and Lou, Qi and Vang, Yeeleng Scott and Xie, Xiaohui},
  booktitle={Medical Image Computing and Computer Assisted Intervention- MICCAI 2017: 20th International Conference, Quebec City, QC, Canada, September 11-13, 2017, Proceedings, Part III 20},
  pages={603--611},
  year={2017},
  organization={Springer}
}

@article{patch_clsf_sliding_wdw2,
  author    = {
                Ridhi Arora and
                Prateek Kumar Rai and
                Balasubramanian Raman},
  title     = {Deep feature-based automatic classification of mammograms},
  journal   = {Med Biol Eng Comput},
  volume    = {58(6)},
  month      = {Jun},
  year      = {2020},
  doi = {10.1007/s11517-020-02150-8}
}

@inproceedings{patch_clsf_sliding_wdw3,
  title={Classification and detection in mammograms with weak supervision via dual branch deep neural net},
  author={Bakalo, Ran and Ben-Ari, Rami and Goldberger, Jacob},
  booktitle={2019 IEEE 16th International Symposium on Biomedical Imaging (ISBI 2019)},
  pages={1905--1909},
  year={2019},
  organization={IEEE}
}

@inproceedings{patch_clsf_sliding_wdw4,
  title={Deformable gabor feature networks for biomedical image classification},
  author={Gong, Xuan and Xia, Xin and Zhu, Wentao and Zhang, Baochang and Doermann, David and Zhuo, Li'an},
  booktitle={Proceedings of the IEEE/CVF Winter Conference on applications of computer vision},
  pages={4004--4012},
  year={2021}
}

@article{ayana2022patchless,
  title={Patchless Multi-Stage Transfer Learning for Improved Mammographic Breast Mass Classification},
  author={Ayana, Getinet and Park, Jong and Choe, Seung Won},
  journal={Cancers},
  volume={14},
  number={5},
  pages={1280},
  year={2022},
  publisher={MDPI},
  doi={10.3390/cancers14051280},
  pmid={35267587},
  pmcid={PMC8909211}
}

@inproceedings{sheller2019multi,
  title={Multi-institutional deep learning modeling without sharing patient data: A feasibility study on brain tumor segmentation},
  author={Sheller, Micah J and Reina, G Anthony and Edwards, Brandon and Martin, Jason and Bakas, Spyridon},
  booktitle={Brainlesion: Glioma, Multiple Sclerosis, Stroke and Traumatic Brain Injuries: 4th International Workshop, BrainLes 2018, Held in Conjunction with MICCAI 2018, Granada, Spain, September 16, 2018, Revised Selected Papers, Part I 4},
  pages={92--104},
  year={2019},
  organization={Springer}
}

@article{2020_FL_breast_density_classification,
   title={Federated Learning for Breast Density Classification: A Real-World Implementation},
   ISSN={1611-3349},
   url={http://dx.doi.org/10.1007/978-3-030-60548-3_18},
   DOI={10.1007/978-3-030-60548-3_18},
   journal={Lecture Notes in Computer Science},
   publisher={Springer International Publishing},
   author={Roth, Holger R. and Chang, Ken and Singh, Praveer and Neumark, Nir and Li, Wenqi and Gupta, Vikash and Gupta, Sharut and Qu, Liangqiong and Ihsani, Alvin and Bizzo, Bernardo C. and et al.},
   year={2020},
   pages={181–191}
}

@article{FL-curriculum-learning-MI,
  title={Memory-aware curriculum federated learning for breast cancer classification},
  author={Jim{\'e}nez-S{\'a}nchez, Amelia and Tardy, Mickael and Ballester, Miguel A Gonz{\'a}lez and Mateus, Diana and Piella, Gemma},
  journal={Computer Methods and Programs in Biomedicine},
  volume={229},
  pages={107318},
  year={2023},
  publisher={Elsevier}
}

@article{konevcny2016federated,
  title={Federated optimization: Distributed machine learning for on-device intelligence},
  author={Kone{\v{c}}n{\`y}, Jakub and McMahan, H Brendan and Ramage, Daniel and Richt{\'a}rik, Peter},
  journal={arXiv preprint arXiv:1610.02527},
  year={2016}
}

@inproceedings{mcmahan2017communication,
  title={Communication-efficient learning of deep networks from decentralized data},
  author={McMahan, Brendan and Moore, Eider and Ramage, Daniel and Hampson, Seth and y Arcas, Blaise Aguera},
  booktitle={Artificial intelligence and statistics},
  pages={1273--1282},
  year={2017},
  organization={PMLR}
}

@article{zhao2018federated,
  title={Federated learning with non-iid data},
  author={Zhao, Yue and Li, Meng and Lai, Liangzhen and Suda, Naveen and Civin, Damon and Chandra, Vikas},
  journal={arXiv preprint arXiv:1806.00582},
  year={2018}
}

@article{rieke2020future,
  title={The future of digital health with federated learning},
  author={Rieke, Nicola and Hancox, Jonny and Li, Wenqi and Milletari, Fausto and Roth, Holger R and Albarqouni, Shadi and Bakas, Spyridon and Galtier, Mathieu N and Landman, Bennett A and Maier-Hein, Klaus and others},
  journal={NPJ digital medicine},
  volume={3},
  number={1},
  pages={1--7},
  year={2020},
  publisher={Nature Publishing Group}
}

@misc{karimireddy2021scaffold,
    title={SCAFFOLD: Stochastic Controlled Averaging for Federated Learning}, 
    author={Sai Praneeth Karimireddy and Satyen Kale and Mehryar Mohri and Sashank J. Reddi and Sebastian U. Stich and Ananda Theertha Suresh},
    year={2021},
    eprint={1910.06378},
    archivePrefix={arXiv},
    primaryClass={cs.LG},
    doi = {10.48550/arXiv.1812.06127},
    url = {https://doi.org/10.48550/arXiv.1812.06127}
}

@misc{li2020fedprox,
    title={Federated Optimization in Heterogeneous Networks}, 
    author={Tian Li and Anit Kumar Sahu and Manzil Zaheer and Maziar Sanjabi and Ameet Talwalkar and Virginia Smith},
    year={2020},
    eprint={1812.06127},
    archivePrefix={arXiv},
    primaryClass={cs.LG},
    doi = {10.48550/arXiv.1812.06127},
    url = {https://doi.org/10.48550/arXiv.1812.06127}
}

@inproceedings{wortsman2022model,
  title={Model soups: averaging weights of multiple fine-tuned models improves accuracy without increasing inference time},
  author={Wortsman, Mitchell and Ilharco, Gabriel and Gadre, Samir Ya and Roelofs, Rebecca and Gontijo-Lopes, Raphael and Morcos, Ari S and Namkoong, Hongseok and Farhadi, Ali and Carmon, Yair and Kornblith, Simon and others},
  booktitle={International conference on machine learning},
  pages={23965--23998},
  year={2022},
  organization={PMLR}
}

@misc{pleiss2017memoryefficientdensenet,
      title={Memory-Efficient Implementation of DenseNets}, 
      author={Geoff Pleiss and Danlu Chen and Gao Huang and Tongcheng Li and Laurens van der Maaten and Kilian Q. Weinberger},
      year={2017},
      eprint={1707.06990},
      archivePrefix={arXiv},
      primaryClass={cs.CV}
}

@article{cbis-ddsm,
  author    = {Lee, Rebecca and Gimenez, Francisco and Hoogi, Assaf and others},
  title     = {A curated mammography data set for use in computer-aided detection and diagnosis research},
  journal   = {Scientific Data},
  volume    = {4},
  pages     = {170177},
  year      = {2017},
  doi       = {10.1038/sdata.2017.177},
  url       = {https://doi.org/10.1038/sdata.2017.177}
}

@article{BreastDtyRace,
    author = {del Carmen, Marcela G. and Halpern, Elkan F. and Kopans, Daniel B. and Moy, Beverly and Moore, Richard H. and Goss, Paul E. and Hughes, Kevin S.},
    title = {Mammographic Breast Density and Race},
    journal = {American Journal of Roentgenology},
    volume = {188},
    number = {4},
    pages = {1147-1150},
    year = {2007},
    doi = {10.2214/AJR.06.0619},
    note ={PMID: 17377060},
    URL = {https://doi.org/10.2214/AJR.06.0619},
    eprint = {https://doi.org/10.2214/AJR.06.0619}
}

@article{cancer_statistics_2020,
author = {Sung, Hyuna and Ferlay, Jacques and Siegel, Rebecca L. and Laversanne, Mathieu and Soerjomataram, Isabelle and Jemal, Ahmedin and Bray, Freddie},
title = {Global Cancer Statistics 2020: GLOBOCAN Estimates of Incidence and Mortality Worldwide for 36 Cancers in 185 Countries},
journal = {CA: A Cancer Journal for Clinicians},
volume = {71},
number = {3},
pages = {209-249},
doi = {https://doi.org/10.3322/caac.21660},
url = {https://acsjournals.onlinelibrary.wiley.com/doi/abs/10.3322/caac.21660},
eprint = {https://acsjournals.onlinelibrary.wiley.com/doi/pdf/10.3322/caac.21660},
year = {2021}
}

@book{BI-RADS,
  author    = {D'Orsi, Carl J. and Sickles, Edward A. and Mendelson, Ellen B. and Morris, Elizabeth A. and others},
  title     = {ACR BI-RADS\textregistered\ Atlas, Breast Imaging Reporting and Data System},
  year      = {2013},
  publisher = {American College of Radiology},
  address   = {Reston, VA}
}

@article{cancer_statistics_2016_miller,
author = {Siegel, Rebecca L. and Miller, Kimberly D. and Jemal, Ahmedin},
title = {Cancer statistics, 2016},
journal = {CA: A Cancer Journal for Clinicians},
volume = {66},
number = {1},
pages = {7-30},
doi = {https://doi.org/10.3322/caac.21332},
url = {https://acsjournals.onlinelibrary.wiley.com/doi/abs/10.3322/caac.21332},
year = {2016}
}

@article{kairouz_FL_review,
  title={Advances and open problems in federated learning},
  author={Kairouz, Peter and McMahan, H Brendan and Avent, Brendan and Bellet, Aur{\'e}lien and Bennis, Mehdi and Bhagoji, Arjun Nitin and Bonawitz, Kallista and Charles, Zachary and Cormode, Graham and Cummings, Rachel and others},
  journal={Foundations and trends{\textregistered} in machine learning},
  volume={14},
  number={1--2},
  pages={1--210},
  year={2021},
  publisher={Now Publishers, Inc.}
}

@article{li2005association,
  title={The association of measured breast tissue characteristics with mammographic density and other risk factors for breast cancer},
  author={Li, Tong and Sun, Limei and Miller, Naomi and Nicklee, Trudey and Woo, Jennifer and Hulse-Smith, Lee and Tsao, Ming-Sound and Khokha, Rama and Martin, Lisa and Boyd, Norman},
  journal={Cancer Epidemiology Biomarkers \& Prevention},
  volume={14},
  number={2},
  pages={343--349},
  year={2005},
  publisher={American Association for Cancer Research}
}

@article{boyd2007mammographic,
  title={Mammographic density and the risk and detection of breast cancer},
  author={Boyd, Norman F and Guo, Helen and Martin, Lisa J and Sun, Limei and Stone, Jennifer and Fishell, Eve and Jong, Roberta A and Hislop, Greg and Chiarelli, Anna and Minkin, Salomon and others},
  journal={New England journal of medicine},
  volume={356},
  number={3},
  pages={227--236},
  year={2007},
  publisher={Mass Medical Soc}
}

@article{ursin2009relative,
  title={The relative importance of genetics and environment on mammographic density},
  author={Ursin, Giske and Lillie, Elizabeth O and Lee, Eunjung and Cockburn, Myles and Schork, Nicholas J and Cozen, Wendy and Parisky, Yuri R and Hamilton, Ann S and Astrahan, Melvin A and Mack, Thomas},
  journal={Cancer Epidemiology Biomarkers \& Prevention},
  volume={18},
  number={1},
  pages={102--112},
  year={2009},
  publisher={American Association for Cancer Research}
}

@article{greendale2003,
    author = {Greendale, Gail A. and Reboussin, Beth A. and Slone, Stacey and Wasilauskas, Carol and Pike, Malcolm C. and Ursin, Giske},
    title = {Postmenopausal Hormone Therapy and Change in Mammographic Density},
    journal = {JNCI: Journal of the National Cancer Institute},
    volume = {95},
    number = {1},
    pages = {30-37},
    year = {2003},
    month = {01},
    issn = {0027-8874},
    doi = {10.1093/jnci/95.1.30},
    url = {https://doi.org/10.1093/jnci/95.1.30},
    eprint = {https://academic.oup.com/jnci/article-pdf/95/1/30/9849351/30.pdf},
}

@article{muthukrishnan2022mammofl,
title={MammoFL: Mammographic Breast Density Estimation using Federated Learning},
author={Muthukrishnan, Ramya and Heyler, Angelina and Katti, Keshava and Pati, Sarthak and Mankowski, Walter and Alahari, Aprupa and Sanborn, Michael and Conant, Emily F and Scott, Christopher and Winham, Stacey and others},
journal={arXiv preprint arXiv:2206.05575},
year={2022}
}

@article{Roth_NVIDIA_FLARE_Federated_2023,
author = {Roth, Holger R. and Cheng, Yan and Wen, Yuhong and Yang, Isaac and Xu, Ziyue and Hsieh, Yuan-Ting and Kersten, Kristopher and Harouni, Ahmed and Zhao, Can and Lu, Kevin and Zhang, Zhihong and Li, Wenqi and Myronenko, Andriy and Yang, Dong and Yang, Sean and Rieke, Nicola and Quraini, Abood and Chen, Chester and Xu, Daguang and Ma, Nic and Dogra, Prerna and Flores, Mona and Feng, Andrew},
doi = {https://doi.org/10.48550/arXiv.2210.13291},
journal = {IEEE Data Eng. Bull., Vol. 46, No. 1},
month = mar,
title = {{NVIDIA FLARE: Federated Learning from Simulation to Real-World}},
year = {2023}
}
\bibliographystyle{spiejour}   % makes bibtex use spiejour.bst

%%%%% Biographies of authors %%%%%

\vspace{2ex}\noindent\textbf{Gonzalo Iñaki Quintana} is a research scientist, focusing on artificial intelligence and deep learning. He holds a PhD in applied mathematics from ENS Paris-Saclay, two engineering degrees in computer science and electrical engineering, and a MSc in applied mathematics. His current research interests include federated learning, contrastive learning, medical imaging, and deep generative models.
\vspace{1ex}

\vspace{2ex}\noindent\textbf{Franco Martin Di Maria} is an engineer specializing in software development and artificial intelligence. He holds engineering degrees in software and computer science from the University of Buenos Aires and IMT Atlantique, and a MSc in image and signal processing from the University of Rennes. He currently works on the development and validation of advanced medical imaging software.
\vspace{1ex}

\noindent Biographies and photographs of the other authors are not available.

\listoffigures
\listoftables

\end{spacing}
\end{document}